\documentclass[]{stacs}
\theoremstyle{plain}
\theoremstyle{definition}

\begin{document}

% \title[short title]{title}
\title[Dichotomy for MinHom]{A dichotomy
theorem for the general minimum cost homomorphism problem}

% \author[ref]{Short author}{Author}
\author[lab]{R. Takhanov}{Rustem Takhanov}
% \address[ref]{Address of authors with ref as reference}
\address[lab]{Department of Computer and Information Science, Link{\"o}ping University}  %required
\email{g-rusta@ida.liu.se}  %optional
\email{takhanov@mail.ru}  %optional

%% mandatory lists of keywords and classifications:

%% the abstract has to PRECEDE the command \maketitle:
%% be sure not to issue the \maketitle command twice!
\sloppy
\begin{abstract}
\noindent In the constraint satisfaction problem ($CSP$), the aim
is to find an assignment of values to a set of variables subject to specified
constraints. In the minimum cost homomorphism problem ($MinHom$), one is additionally given
weights $c_{va}$ for every variable $v$ and value $a$, and the aim is to
find an assignment $f$ to the variables that minimizes $\sum_{v}
c_{vf(v)}$. Let $MinHom\left( \Gamma \right)$ denote the $MinHom$ problem
parameterized by the set of predicates allowed for constraints.
$MinHom\left( \Gamma \right)$ is related to many well-studied
combinatorial optimization problems, and concrete applications can be found in, for instance,
defence logistics and machine learning. We show that $MinHom\left( \Gamma \right)$ can be studied
by using algebraic methods similar to those used for CSPs.
With the aid of algebraic techniques, we classify the
computational complexity of $MinHom\left( \Gamma \right)$ for all choices of $\Gamma$.
Our result settles a general dichotomy conjecture previously resolved only for certain
classes of directed graphs, [Gutin, Hell, Rafiey, Yeo, European J. of Combinatorics, 2008].
\end{abstract}

\maketitle

%% start the paper here:
\section{Introduction}\label{S:one}

Constraint satisfaction problems ($CSP$) are a natural way of formalizing a large number of
computational problems arising in combinatorial optimization, artificial intelligence,
and database theory. This problem has the following two equivalent formulations:
(1) to find an assignment of values to a given set of variables, subject to constraints on the values that can be
assigned simultaneously to specified subsets of variables, and
(2) to find a homomorphism between two finite relational structures $A$ and $B$.
Applications of $CSP$s arise in the propositional logic, database and graph theory, scheduling and many other areas.
During the past 30 years, $CSP$ and its subproblems has been intensively studied by computer scientists and mathematicians.
Considerable attention has been given to the case where the
constraints are restricted to a given finite set of relations $\Gamma$, called a constraint language \cite{bulatov,feder,jeavons,schaefer}.
For example, when $\Gamma$ is a constraint language over the boolean set $\{0, 1\}$ with four ternary predicates $x\vee y\vee z$,
$\overline{x}\vee y\vee z$, $\overline{x}\vee \overline{y}\vee z$,
$\overline{x}\vee \overline{y}\vee \overline{z}$ we obtain 3-SAT.
This direction of research has been mainly concerned with the computational complexity of $CSP\left( \Gamma \right)$ as a function of $\Gamma$.
It has been shown that the complexity of $CSP\left( \Gamma \right)$ is highly connected with relational clones of universal algebra \cite{jeavons}. For every constraint language $\Gamma$, it has been conjectured that $CSP\left( \Gamma \right)$ is either in P or NP-complete \cite{feder}.

In the minimum cost homomorphism problem ($MinHom$), we are given variables subject to constraints and, additionally, costs on variable/value pairs. Now, the task is not just to find any satisfying assignment to the variables, but one that minimizes the total cost.

\begin{defi}\label{D:first}
Suppose we are given a finite domain set $A$ and a finite constraint language $\Gamma \subseteq \bigcup\limits_{k = 1}^\infty  {{2^{{A^k}}}} $.
Denote by
$MinHom\left( \Gamma \right)$
the following minimization task:

\noindent {\bf Instance:} A first-order formula
$\Phi \left( {{x_1},\dots,{x_n}} \right) =
\mathop  \wedge \limits_{i = 1}^N {\rho _i}\left( {{y_{i1}},\dots,{y_{i{n_i}}}} \right)$,
${\rho _i} \in \Gamma,{y_{ij}} \in \left\{ {{x_1},\dots,{x_n}} \right\}$, and
weights ${w_{ia}}\in {\mathbb N}, 1 \leq i \leq n, a \in A$.

\noindent {\bf Solution:} Assignment $f:\left\{ {{x_1},\dots,{x_n}} \right\} \rightarrow A$, that satisfies
the formula $\Phi $. If there is no such assignment, then indicate it.

\noindent {\bf Measure:} $\sum\limits_{i = 1}^n {w_{if\left( x_{i} \right)}}$.
\end{defi}

\begin{rem}\label{R:0}
Note that when we require weights to be positive we do not lose generality,
since $MinHom\left( \Gamma \right)$ with arbitrary weights can be polynomial-time reduced to $MinHom\left( \Gamma \right)$ with positive weights by the following trick:
we can add $s$ to all weights, where
$s$ is some integer. This trick only adds $ns$ to the value of the optimized measure.
Hence, we can make all weights negative, and $MinHom\left( \Gamma \right)$ modified this way is
equivalent to maximization but with positive weights only.
This remark explains why both names $MinHom$ and $MaxHom$ can be allowed, though we prefer
$MinHom$ due to historical reasons.
\end{rem}

$MinHom$ was introduced in \cite{gutin2} where it was motivated by
a real-world problem in defence logistics. The question for which directed graphs $H$ the problem
$MinHom\left( \left\{ H \right\} \right)$ is polynomial-time solvable was considered in \cite{gupta,gutin0,gutin1,gutin2,gutin3}. In this paper, we approach the problem in its most general form by algebraic methods and
give a complete algebraic characterization of tractable constraint languages.
From this characterization, we obtain a dichotomy for $MinHom$,
i.e., if $MinHom\left( \Gamma \right)$ is not polynomial-time solvable, then it is NP-hard.
Of course, this dichotomy implies the dichotomy for directed graphs.

In Section 2, we present some preliminaries together
with results connecting the complexity of $MinHom$ with
conservative algebras. The main dichotomy theorem is
stated in Section 3 and its proof is divided into
several parts which can be found in Sections 4-8.
The NP-hardness results are collected in Section 4 followed
by the building blocks for the tractability result: existence of
majority polymorphisms (Section 5) and
connections with optimization in perfect graphs (Section 6).
Section 7 introduces the concept of {\em arithmetical deadlocks}
which lay the foundation for the final proof in Section 8.
In Section 9 we reformulate our main result in terms of relational clones.
Finally, in Section 10 we explain the relation of our results to previous research and
present directions for future research.

\section{Algebraic structure of tractable constraint languages}

Recall that an
optimization problem $A$ is called NP-hard if some NP-complete language can be
recognized in polynomial time with the aid of an oracle for $A$. We assume that $P \ne NP$.

\begin{defi}\label{D:first}
Suppose we are given a finite set $A$ and a constraint language
$\Gamma \subseteq \bigcup\limits_{k = 1}^\infty  {{2^{{A^k}}}} $. The language $\Gamma$ is
said to be {\em tractable} if, for every finite subset
$\Gamma' \subseteq \Gamma$, $MinHom\left( \Gamma' \right)$ is polynomial-time solvable, and $\Gamma$ is
called {\em NP-hard} if there is a finite subset $\Gamma' \subseteq \Gamma$ such that
$MinHom\left( \Gamma' \right)$ is NP-hard.
\end{defi}

First, we will state some standard definitions from universal algebra.

\begin{defi}\label{D:first}
Let $\rho  \subseteq A^m $ and $f:A^n  \to A$.
We say that the function (operation) $f$ {\em preserves} the predicate $\rho $ if,
for every $\left( {x_1^i ,\dots,x_m^i } \right) \in \rho, 1 \leq i \leq n$, we have that $\left( {f\left( {x_1^1 ,\dots,x_1^n
} \right),\dots,f\left( {x_m^1 ,\dots,x_m^n } \right)} \right) \in
\rho $.
\end{defi}

For a constraint language $\Gamma$, let $Pol\left( \Gamma \right)$ denote the
set of operations preserving all predicates in $\Gamma$. Throughout the paper, we let $A$ denote a finite domain and $\Gamma$ a constraint language over $A$. We assume the domain $A$ to be finite.

\begin{defi}\label{D:first}
A constraint language $\Gamma$ is called {\em a relational
clone} if it contains every predicate expressible by a first-order formula involving only
\begin{itemize}
    \item predicates from $\Gamma \cup \left\{=^{A}\right\}$;
    \item conjunction; and
    \item existential quantification.
\end{itemize}
\end{defi}

First-order formulas involving only conjunction and existential quantification are
often called {\em primitive positive (pp) formulas}. For a given constraint language $\Gamma$, the set of all predicates that can be
described by pp-formulas over $\Gamma$ is called the {\em closure} of $\Gamma$ and is denoted by $\langle \Gamma \rangle$.

For a set of
operations $F$ on $A$, let $Inv\left( F \right)$ denote the set of
predicates preserved under the operations of $F$. Obviously, $Inv\left( F \right)$ is a relational clone.
The next result is well-known \cite{bodnarchuk,geiger}.

\begin{thm}\label{T:1}
For a constraint language $\Gamma$ over a finite set $A$, $\langle \Gamma \rangle = Inv\left( Pol\left( \Gamma \right) \right)$.
\end{thm}

Theorem \ref{T:1} tells us that the Galois closure of a constraint language $\Gamma$
is equal to the set of all predicates that can be obtained
via pp-formulas from the predicates in $\Gamma$.

\begin{thm}\label{T:2}
For any finite constraint language $\Gamma$ and any finite $\Gamma'\subseteq \langle \Gamma \rangle$,
there is a polynomial time reduction from $MinHom\left( \Gamma' \right)$ to $MinHom\left( \Gamma \right)$.
\end{thm}

\proof[Proof.]
Since any predicate from $\Gamma'$ can be viewed as a
pp-formula with predicates in $\Gamma$, an input formula to $MinHom\left( \Gamma' \right)$
can be represented on the form $\Phi \left( {{x_1},\dots,{x_n}} \right) =
\mathop  \wedge \limits_{i = 1}^N \exists {z_{i1}},\dots,{z_{i{m_i}}}{\rm{ }}
{\Phi _i}\left( {{y_{i1}},\dots,{y_{i{n_i}}},{z_{i1}},\dots,{z_{i{m_i}}}} \right)$, where ${y_{ij}} \in \left\{ {{x_1},\dots,{x_n}} \right\}$ and
$\Phi _i$ is a first-order formula involving only predicates in $\Gamma$, equality, and conjunction.
Obviously, this formula is equivalent to $\exists {z_{11}},\dots,{z_{N{m_N}}}\mathop  \wedge \limits_{i = 1}^N {\Phi _i}\left( {{y_{i1}},\dots,{y_{i{n_i}}},{z_{i1}},\dots,{z_{i{m_i}}}} \right)$. $\mathop\wedge \limits_{i = 1}^N {\Phi _i}\left( {{y_{i1}},\dots,{y_{i{n_i}}},{z_{i1}},\dots,{z_{i{m_i}}}} \right)$
can be considered as an instance of $MinHom\left( \Gamma \cup \left\{=^{A}\right\} \right)$ with variables ${x_1},\dots,{x_n}, {z_{11}},\dots,{z_{N{m_N}}}$ where weights
$w_{ij}$ will remain the same and for additional variables $z_{kl}$ we define $w_{z_{kl}j} = 0$.
By solving $MinHom\left( \Gamma \cup \left\{=^{A}\right\} \right)$
with the described input, we can find a solution of the initial $MinHom\left( \Gamma' \right)$ problem. It is easy to see that the number of added variables is bounded by a polynomial in $n$. So this reduction can be carried out in polynomial time.
Finally, $MinHom\left( \Gamma \cup \left\{=^{A}\right\} \right)$ can be reduced polynomially to $MinHom\left( \Gamma \right)$ because an equality constraint for a pair of variables is equivalent to identification of these variables.
\qed

The previous theorem tells us that the complexity of $MinHom\left( \Gamma \right)$ is basically
determined by $Inv\left( Pol\left( \Gamma \right) \right)$, i.e., by $Pol\left( \Gamma \right)$.
That is why we will be concerned with the classification of sets of operations $F$ for which $Inv\left( {F} \right)$ is a tractable constraint language.

\begin{defi}\label{D:first}
An {\em algebra} is an ordered pair ${\mathbb A} = \left(A, F\right)$ such that $A$ is
a nonempty set (called a universe) and $F$ is a family of finitary operations on $A$. An
algebra with a finite universe is referred to as a finite algebra.
\end{defi}

\begin{defi}\label{D:first}
An algebra ${\mathbb A} = \left(A, F\right)$ is called {\em tractable} if $Inv(F)$ is a tractable constraint language and ${\mathbb A}$ is called {\em NP-hard} if $Inv(F)$ is an NP-hard constraint language.
\end{defi}

In the following theorem, we
show that we only need to consider a very special type of algebras, so called {\em conservative} algebras.

\begin{defi}\label{D:first}
An algebra ${\mathbb A} = \left(A, F\right)$ is called conservative if for every operation $f \in F$ we have that $f\left( {x_1 ,\dots,x_n } \right)
\in \left\{ {x_1 ,\dots,x_n } \right\}$.
\end{defi}

\begin{thm}\label{T:3}
For any finite constraint language $\Gamma$ over $A$ and $C \subseteq A$,
there is a polynomial time Turing reduction from $MinHom\left( \Gamma \cup \left\{ C \right\} \right)$ to $MinHom\left( \Gamma \right)$.
\end{thm}

\proof[Proof.]
Let the first-order formula
$\Phi \left( {{x_1},\dots,{x_n}} \right) = \mathop  \wedge \limits_{i = 1}^{M} C\left(y_i\right)
\wedge
\mathop  \wedge \limits_{i = 1}^N
{\rho _i}\left( {{z_{i1}},\dots,{z_{i{n_i}}}} \right)$, where $\rho _i \in \Gamma, y_i, {z_{ij}} \in \left\{ {{x_1},\dots,{x_n}} \right\}$,
and weights ${w_{ia}},1\leq i\leq n ,a \in A$ be an instance of $MinHom\left( \Gamma \cup \left\{ C \right\} \right)$. We assume without loss of generality that $y_i\ne y_j$, when $i\ne j$.
Let $W = \sum\limits_{i =
1}^n\sum\limits_{a\in A} {w_{ia}}  + 1$ and define a new formula and weights
\[
\Phi' \left( {{x_1},\dots,{x_n}} \right) =
\mathop  \wedge \limits_{i = 1}^N
{\rho _i}\left( {{z_{i1}},\dots,{z_{i{n_i}}}} \right)
\]
\[
{w'_{ia}} = \left\{ {\begin{array}{*{20}{c}}
   {{w_{ia}} + W,{\rm{if\,\,}}a \notin C,\exists j{\rm{\,\,}}{x_i} = {y_j}}  \\
   {{w_{ia}},{\rm{\,\,otherwise\,\,\,\,\,\,\,\,\,\,\,\,\,\,\,\,\,\,\,\,\,\,\,\,\,\,\,\,\,\,\,\,\,\,\,\,\,}}}  \\
\end{array}} \right.
\]

Then, using an oracle for $MinHom\left( \Gamma \right)$, we can solve
\[
\mathop { \min }\limits_{f\rm{\,\,satisfies\,\,}\Phi'}
\sum\limits_j {w'_{jf\left(x_j\right)} }.
\]
Suppose that $\Phi \left( {{x_1},\dots,{x_n}} \right)$ is satisfiable and $f$ is a satisfying assignment. It is easy to see that
the part of the measure $\sum\limits_j {w'_{jf\left(x_j\right)} }$ that corresponds to the added values $W$ is equal to 0 and the measure cannot be greater than $W - 1$. If $g$ is any assignment that does not satisfy $\mathop  \wedge \limits_{i = 1}^{M} C\left(y_i\right)$, then we see that this part of measure cannot be 0, and hence, is greater or equal to $W$. This means that the minimum in the task is achieved on satisfying assignments of $\Phi \left( {{x_1},\dots,{x_n}} \right)$ and any such assignment minimize the part of the measure that corresponds to the initial weights, i.e., $\sum\limits_i {w_{if\left(x_i\right)} }$.

If $\Phi \left( {{x_1},\dots,{x_n}} \right)$ is not satisfiable, then either $\Phi'$ is not satisfiable or $\mathop { \min }\limits_{f\rm{\,\,satisfies\,\,}\Phi'}
\sum\limits_j {w'_{jf\left(x_j\right)} } \geq W$. Using an oracle for $MinHom\left( \Gamma \right)$, we can easily check this.

Consequently, $MinHom\left( \Gamma \cup \left\{ C \right\} \right)$ is polynomial-time
reducible to $MinHom\left( \Gamma \right)$.
\qed

\begin{thm}\label{T:4}
If $\Gamma$ is a constraint language over $A$ that contains all unary relations, then ${\mathbb A} = \left(A, Pol\left( \Gamma \right)\right)$ is conservative.
\end{thm}

\proof[Proof.]
Let $C = \left\{ {x_1 ,\dots,x_n } \right\} \subseteq
A$. If a function $f:A^n\rightarrow A$ preserves the predicate $C$, then
$f\left( {x_1 ,\dots,x_n } \right) \in \left\{ {x_1 ,\dots,x_n }
\right\}$.
\qed

\section{Structure of tractable conservative algebras}

Let $g:A^k  \to A$ be an arbitrary conservative function and $S \subseteq A$. Define the function
$g|_S :S^k  \to S$, such that $\forall x_1 ,\dots,x_k  \in S{\rm{\,\,}}g|_S \left( {x_1 ,\dots,x_k } \right) = g\left( {x_1 ,\dots,x_k } \right)$, i.e. the restriction of $g$ to the set $S$. Throughout this paper we will consider a conservative algebra ${\mathbb A} = \left(A, F\right)$. For every $B \subseteq A$, let $F|_B  = \left\{ {f_B |f \in F} \right\}$. Then ${\mathbb A}|_{B}$ denotes an algebra $\left(B, F_{B}\right)$.
We assume that $F$ is closed under superposition and variable change and contains all projections, i.e., it is {\em a functional clone}, because closing the set $F$ under these operations does not change the set $Inv\left( {F} \right)$.

Sometimes we will consider clones as algebras and to describe them we will use the terms (conservativeness, tractability, NP-hardness) defined for algebras.
All tractable clones, in case $A = \{0,1\}$, can be easily found using well-known classification of boolean clones \cite{post}.

\begin{thm}\label{T:PNew}
The boolean functional clone $H$ is tractable if either $\left\{x\wedge y, x\vee y\right\} \subseteq H$ or $\left\{\left( {x \wedge \overline y } \right) \vee \left( {\overline y  \wedge z} \right) \vee \left( {x \wedge z} \right)\right\} \subseteq H$, where $\wedge, \vee$ denote conjunction and disjunction.
Otherwise, $H$ is NP-hard.
\end{thm}

In the proof of this theorem we will need the following definition.

\begin{defi}\label{D:first}
A constraint language $\Gamma$ over $A$ is called a
{\em maximal tractable} constraint language if it is tractable and is not
contained in any other tractable languages.
\end{defi}

Let us identify all maximal tractable constraint languages in the boolean case using Post`s
classification \cite{post}.
From Theorems \ref{T:2}, \ref{T:3}, \ref{T:4} we conclude that every maximal tractable constraint language corresponds to some conservative
functional clone.
In the case $A = \left\{ {0,1} \right\}$, there is a countable number
of conservative clones: we list them below according to the table on page 76 \cite{marchenkov}. For every row, the closure of the
predicates given is equal to the set of all predicates preserved
under the functions of the corresponding clone.
\[
\begin{array}{*{20}c}
   {T_{01} } & {x = 0,x = 1}  \\
   {M_{01} } & {x = 0,x = 1,x_1  \le x_2 }  \\
   {S_{01} } & {x = 0,x_1  \ne x_2 }  \\
   {SM} & {x_1  \ne x_2 ,x_1  \le x_2 }  \\
   {L_{01} } & {x = 1,x_1  \oplus x_2  \oplus x_3  = 0}  \\
   {U_{01} } & {x = 0,x = 1,x_1  = x_2  \vee x_1  = x_3 }  \\
   {K_{01} } & {x = 0,x = 1,x_1  = x_2 x_3 }  \\
   {D_{01} } & {x = 0,x = 1,x_1  = x_2  \vee x_3 }  \\
   {I_1^m } & {x = 1,x_1 x_2 \dots x_m  = 0}  \\
   {MI_1^m } & {x = 1,x_1  \le x_2 ,x_1 x_2 \dots x_m  = 0}  \\
   {O_0^m } & {x = 0,x_1  \vee x_2  \vee \dots \vee x_m  = 1}  \\
   {MO_0^m } & {x = 0,x_1  \le x_2 ,x_1  \vee x_2  \vee \dots \vee x_m  = 1}  \\
\end{array}
\]
where $x\oplus y = x+y\left(mod\rm{\,\,}2\right)$.

\begin{lem}\label{L:Post}
The relational clones
$Inv\left( {M_{01} } \right)$ and $Inv\left( {S_{01} } \right)$
are maximal tractable constraint languages. Every other constraint language given in the table, except $Inv\left( {T_{01} } \right)$,
is NP-hard.
\end{lem}

\proof[Proof.]
The class $Inv\left( {T_{01} } \right)$ is tractable, since it
contains only two simple unary predicates
$\{0\}$ and $\{1\}$. As we will see later, it cannot be
maximal since it is included in other tractable constraint languages.

Let us prove that $Inv\left( {M_{01} }
\right)$ is tractable. By Theorem \ref{T:2}, it is
equivalent to polynomial solvability of $MinHom\left( {\left\{
{\left\{ 0 \right\},\left\{ 1 \right\},\left\{ {\left( {x_1 ,x_2
} \right)|x_1  \le x_2 } \right\}} \right\}} \right)$, because the
class $Inv\left( {M_{01} } \right)$ is the closure of this set of
predicates. A proof of this statement can be found in \cite{jonnson}.
We will give it for completeness.

Obviously, $MinHom\left( {\left\{
{\left\{ 0 \right\},\left\{ 1 \right\},\left\{ {\left( {x_1 ,x_2
} \right)|x_1  \le x_2 } \right\}} \right\}} \right)$ is equivalent to the following
boolean linear programming task, sets $Q_0,Q_1 \subseteq \left\{1,\dots,n\right\}, Q \subseteq \left\{1,\dots,n\right\}^2$ and
integer weights $w_1,\dots,w_n$ given as an input:
\[
\left\{ {\begin{array}{*{20}{c}}
   {\min\sum\limits_i {{w_i}{x_i}} }  \\
   {{x_i} = 0,i \in {Q_0}}  \\
   {{x_i} = 1,i \in {Q_1}}  \\
   {{x_i} \le {x_j},\left( {i,j} \right) \in Q}  \\
   {{x_i} \in \left\{ {0,1} \right\}} \\
\end{array}} \right.
\]

Let us prove that the polyhedron which is given by the same equalities and inequalities as previous, but
with ${x_i} \in \left\{ {0,1} \right\}$ replaced by $0 \leq {x_i} \leq 1$, is integer. Suppose it is not integer and
$v = ||v_1, v_2, \dots, v_n||^{T}$ is its extreme point where $v_r$ is not equal to 0 or 1.
Let us define $\epsilon$ as the minimum of three values $\mathop {\min}\limits_{v_i \ne v_j} |v_i - v_j|$,
$\mathop {\min}\limits_{v_i \ne 0} |v_i|$, $\mathop {\min}\limits_{v_i \ne 1} |1 - v_i|$ and two vectors $v^{+}$ and $v^{-}$:
$v^{+}_i = v^{-}_i = v_i$ if $v_i \ne v_r$ and
$v^{+}_i = v_i + \epsilon$, $v^{-}_i = v_i - \epsilon$, otherwise. It is easy to see that
points $v^{+}$ and $v^{-}$ are also in polyhedron, and $v = \frac{v^{+} + v^{-}}{2}$. This
contradicts the extremeness of $v$.

Since the polyhedron is integer we can solve $MinHom\left( {\left\{
{\left\{ 0 \right\},\left\{ 1 \right\},\left\{ {\left( {x_1 ,x_2
} \right)|x_1  \le x_2 } \right\}} \right\}} \right)$ in polynomial time by standard linear programming algorithms.
Consequently, $Inv\left(
{M_{01} } \right)$ is tractable.

Now let us prove that $Inv\left( {S_{01} } \right)$ is tractable, i.e. $MinHom\left( \left\{ \left\{ 0
\right\},\left\{ \left( {x_1 ,x_2 } \right)|x_1  \ne x_2
\right\}\right\} \right)$ is polynomial-time solvable.

Let an instance of this problem be the sets $Q_0\subseteq \left\{1,\dots,n\right\}, Q \subseteq \left\{1,\dots,n\right\}^2$ and
integer weights $w_{10},\dots,w_{n0}, w_{11},\dots,w_{n1}$. By $\Phi\left( {Q_0,Q} \right)$ we denote the set of assignments of variables $x_1,\dots,x_n$ that satisfy the input formula, i.e. such that $x_i = 0, i\in Q_0$ and $x_k \ne x_l, (k,l)\in Q$.

The graph $\left( {\left\{1,\dots,n\right\},Q'} \right)$ where $Q' = \{(x,y)|(x,y)\in Q \vee (y,x)\in Q\}$ can be decomposed into connected components $\left( {\left\{1,\dots,n\right\},Q'} \right) =
K_1  \cup \dots \cup K_t $, where $K_i  = \left( {V_i ,E_i }
\right)$. Such a decomposition can be made in $O\left( {n^2 } \right)$ steps. If among these components there is a
graph with an odd cycle, then, obviously, $\Phi\left( {Q_0,Q} \right)
= \emptyset $. Otherwise, the optimization task can be reduced to
subtasks for every component. I.e., if for some component $\Phi\left( {Q_0\cap V_i, E_i} \right) = \emptyset$, then $\Phi\left( {Q_0,Q} \right)
= \emptyset $, otherwise:
\[
\mathop {\min }\limits_{\overline{x} \in \Phi\left( {Q_0,Q} \right)}
\sum\limits_{i = 1}^n {w_{ix_i}}  = \sum\limits_{i = 1}^t {\mathop {\min }\limits_{\overline{x} \in \Phi\left( {Q_0\cap V_i, E_i} \right)} \sum\limits_{j \in V_i } {w_{jx_j}} }.
\]
But $\left| {\Phi\left( {Q_0\cap V_i, E_i} \right)} \right| \le 2$, and a straightforward
algorithm solves every subtask. So, $Inv\left( {S_{01} } \right)$
is tractable.

We first now show that the classes in the table, except $Inv\left( {M_{01} } \right)$,
$Inv\left( {S_{01} } \right)$ and $Inv\left( {T_{01} } \right)$,
are NP-hard.
Since,
\[
\begin{array}{l}
 x_1  \vee x_2  = \exists x_3 \left[ {x_1  \ne x_3 } \right]\wedge \left[ {x_3  \le x_2 } \right] \\
 x_1  \vee x_2  = \exists x_3 \left[ {x_3  = 1} \right]\wedge \left[ {x_3  = x_1  \vee x_3  = x_2 } \right] \\
 \overline {x_1 }  \vee \overline {x_2 }  = \exists x_3 \left[ {x_3  = 0} \right]\wedge \left[ {x_3  = x_1 x_2 } \right] \\
 x_1  \vee x_2  = \exists x_3 \left[ {x_3  = 1} \right]\wedge \left[ {x_3  = x_1  \vee x_2 } \right] \\
 \overline {x_1 }  \vee \overline {x_2 }  = \exists x_3 \dots x_m \left[ {x_1 x_2 \dots x_m  = 0} \right]\wedge \left[ {x_2  = x_3 } \right]\wedge \dots\wedge \left[ {x_{m - 1}  = x_m } \right] \\
 x_1  \vee x_2  = \exists x_3 \dots x_m \left[ {x_1  \vee x_2  \vee \dots \vee x_m  = 1} \right]\wedge \left[ {x_2  = x_3 } \right]\wedge \dots\wedge \left[ {x_{m - 1}  = x_m } \right] \\
 \end{array}
\]
we see that $\left\{ {\left( {x_1 ,x_2 } \right)|x_1  \vee x_2 }
\right\} \in Inv\left( {SM} \right)$, $Inv\left( {U_{01} }
\right),Inv\left( {D_{01} } \right)$, $Inv\left( {O_0^m }
\right),Inv\left( {MO_0^m } \right)$ and $\left\{ {\left( {x_1
,x_2 } \right)|\overline {x_1 }  \vee \overline {x_2 } } \right\}
\in Inv\left( {K_{01} } \right)$, $Inv\left( {I_1^m } \right)$, $
Inv\left( {MI_1^m } \right)$.

We first prove that
$MinHom\left( {\left\{ {\left\{
{\left( {x_1 ,x_2 } \right)|x_1  \vee x_2 } \right\}} \right\}}
\right)$ is NP-hard. Suppose an instance of this problem consists of an undirected graph $G=\left(V,E\right)$ where each vertex is considered as a variable. For each pair of variables $(u,v)\in E$, we require their assignments to satisfy $u = 1$ or $v = 1$. It is easy to see
that for any such assignment $f$, the set $\{x|f(x)=0\}$ is independent
in the graph $G$. Furthermore, for any independent set $S$ in the graph $G$,
$g(x) = [x\notin S]$ is a satisfying assignment.
If we define $w_{i0} = 1, w_{i1} = 1$ for $i\in V$, then
$MinHom$ is equivalent to finding a maximum independent set.
This implies that $MinHom\left( {\left\{ {\left\{
{\left( {x_1 ,x_2 } \right)|x_1  \vee x_2 } \right\}} \right\}}
\right)$ is NP-hard, since finding independent sets of maximal size is an NP-hard problem. The case $MinHom\left( {\left\{ {\left\{ {\left( {x_1 ,x_2 }
\right)|\overline {x_1 }  \vee \overline {x_2 } } \right\}}
\right\}} \right)$ is analogous.

Therefore, $Inv\left( {SM} \right)$,
$Inv\left( {U_{01} } \right)$, $Inv\left( {D_{01} } \right)$,
$Inv\left( {O_0^m } \right)$, $Inv\left( {MO_0^m } \right)$,
$Inv\left( {K_{01} } \right)$, $Inv\left( {I_1^m } \right)$,
$Inv\left( {MI_1^m } \right)$ are NP-hard, too.

It remains to prove NP-hardness of $Inv\left( {L_{01} } \right)$.
We show that using an algorithm for $MinHom\left( {\left\{ {\left( {x_1 ,x_2 ,x_3 } \right)|x_1  \oplus x_2 \oplus x_3  = 1}
\right\}} \right)$ as an oracle, we can solve Max-CUT in
polynomial time.

Let $G = \left( {V,E} \right)$ be a graph and introduce
variables $x_{ij} ,y_i ,y_j ,i,j \in V$. A system of equations $x_{ij} \oplus y_i  \oplus y_j  = 1,i,j \in V$ can be viewed as an
instance of $MinHom\left( {\left\{ {\left( {x_1 ,x_2 ,x_3 } \right)|x_1  \oplus x_2 \oplus x_3  = 1}
\right\}} \right)$. It is easy to see that arbitrary boolean vector
$\overline y = \left( {y_1 ,\dots,y_{\left| V \right|} } \right)$ defines a single solution $x_{ij} = y_i \oplus y_j  \oplus 1,i,j \in V$ of the system. Vector $\overline y $ can be considered as the cut
$\left\{ {i|y_i  = 1} \right\} \subseteq V$ and the value $\sum\limits_{ij}(1-x_{ij})$ is equal to the doubled cost of the cut. Then Max-CUT is polynomially reduced to solving $MinHom\left(Inv\left( {L_{01} } \right) \right)$.

Only two classes $Inv\left( {M_{01} } \right)$ and
$Inv\left( {S_{01} } \right)$ are left as candidates for maximality. Since they are not
included in each other, they are both maximal.

\qed

\begin{lem}\label{L:Post2}
If a constraint language $S \subseteq \bigcup\limits_{k = 1}^\infty  {{2^{{\left\{0,1\right\}^k}}}} $ is contained in neither $Inv\left( {M_{01} } \right)$ nor $Inv\left( {S_{01} } \right)$, then it is NP-hard.
\end{lem}

\proof[Proof.] Suppose we are given a constraint language $S$ which is not contained in $Inv\left( {M_{01} } \right)$ and $Inv\left( {S_{01} } \right)$.
Then, $\langle S\cup 2^{A}\rangle$ is not contained in $Inv\left( {M_{01} } \right)$ and $Inv\left( {S_{01} } \right)$, either. Since $\langle S\cup 2^{A}\rangle$ is a boolean conservative relational clone, then, by previous lemma, it is NP-hard. By Theorems \ref{T:2} and \ref{T:3}, we conclude that $S$ is NP-hard.
\qed

\proof[Proof of Theorem \ref{T:PNew}.]
The bases in the clones $M_{01}, S_{01}$ are $\left\{ { \wedge , \vee } \right\}$ and $\left\{ {\left( {x \wedge \overline y } \right) \vee \left( {\overline y  \wedge z} \right) \vee \left( {x \wedge z} \right)} \right\}$ and the theorem follows from Lemma \ref{L:Post2}.

\qed

Every 2-element subalgebra of a tractable algebra must be tractable, which motivates the following definition.

\begin{defi}\label{D:first}
Let $F$ be a conservative functional clone. We say that $F$ satisfies the {\em necessary local conditions} if and only if for every 2-element subset $B\subseteq A$, either
\begin{itemize}
    \item there exists $f^\wedge, f^\vee\in F$ s.t. $f^\wedge|_B$ and $f^\vee|_B $ are different binary commutative functions; or
    \item there exists $f\in F$ s.t. $f|_B \left( {x,x,y} \right) = f|_B \left( {y,x,x} \right) = f|_B \left( {y,x,y} \right) = y$.
\end{itemize}
\end{defi}

\begin{thm}\label{T:6}
Suppose $F$ is a conservative functional clone. If $F$ is tractable, then it satisfies the necessary local conditions. If $F$ does not satisfy the necessary local conditions, then it is NP-hard.
\end{thm}

\proof[Proof.]
Since for every two-element subset $B \subseteq A$, $Inv\left( {F|_B } \right) \subseteq Inv\left( F \right)$, then $F|_B $ is
tractable. Assume without loss of generality that $B = \{0,1\}$.
From Theorem \ref{T:PNew}, we get that $\left\{ { \wedge , \vee } \right\} \subseteq F|_B $ or $\left\{ {a\left( {x,y,z} \right) = \left( {x \wedge \overline y } \right) \vee \left( {\overline y  \wedge z} \right) \vee \left( {x \wedge z} \right)} \right\} \subseteq F|_B $.
$ \wedge , \vee $ is a pair of different commutative conservative functions and $a \left( {x,x,y} \right) = a \left( {y,x,x} \right) = a \left( {y,x,y} \right) = y$.
\qed

In general, the necessary local conditions are not sufficient for
tractability of a conservative clone. Let $M = \left\{ {B|B \subseteq A, \left| B \right| = 2,F|_B {\rm{\,\,contains\,\,different\,\,binary\,\,commutative\,\,functions}}} \right\}$
and $\overline M  = \left\{ {B|B \subseteq A, \left| B \right| = 2} \right\}\backslash M$.

Suppose $f\in F$. By $\mathop  \downarrow \limits_b^a f$ we mean $a \ne b$ and $f \left( {a,b} \right) = f \left( {b,a} \right) = b$.
For example, $\mathop  \downarrow \limits_2^1 \mathop  \downarrow \limits_3^2 \mathop  \downarrow \limits_3^1 f $
means that $f |_{\left\{ {1,2,3} \right\}} \left( {x,y} \right) = \max \left( {x,y} \right)$.

Introduce an undirected graph without loops $T_F = \left( {M^o ,P} \right)$ where $M^o  = \left\{ {\left( {a,b} \right)|\left\{ {a,b} \right\} \in M} \right\}$ and $P = \left\{ {\left\langle {\left( {a,b} \right),\left( {c,d} \right)} \right\rangle |\left( {a,b} \right),\left( {c,d} \right) \in M^o ,{\rm{\,\,there\,\,is\,\,no\,\,}}f  \in F:\mathop  \downarrow \limits_b^a \mathop  \downarrow \limits_d^c f } \right\}$. The core result of the paper is the following.

\begin{thm}\label{T:7}
Suppose $F$ satisfy the necessary local conditions. If the graph $T_F = \left( {M^o ,P} \right)$ is bipartite, then $F$
is tractable. Otherwise, $F$ is NP-hard.
\end{thm}

The proof of this theorem will be given in two steps. Firstly, in the following section, we will prove NP-hardness of $F$ when $T_F = \left( {M^o ,P} \right)$ is not bipartite. The final sections will be dedicated to the polynomial-time solvable cases.

\section{NP-hard case}
In this section, we will prove that if a set of functions $F$ satisfies the necessary local conditions and $T_F = \left( {M^o ,P} \right)$ (as defined in the previous section) is not bipartite, then $F$ is NP-hard. Let  $\begin{picture}(10,10)\end{picture}_b^a\begin{picture}(10,10)\put(0,10){\line(1,-1){10}}\put(0,0){\line(1,1){10}}\put(0,10){\line(1,0){10}}\end{picture}^{c}_{d} $ and
$\begin{picture}(10,10)\end{picture}_b^a\begin{picture}(10,10)\put(0,10){\line(1,-1){10}}\put(0,0){\line(1,1){10}}\end{picture}^{c}_{d}$
denote the predicates $\left\{ {a,b} \right\} \times \left\{ {c,d} \right\}\backslash \left\{ {\left( {b,d} \right)} \right\}$
and $\left\{ {\left( {a,d} \right),\left( {b,c} \right)} \right\}$, where $a \ne b,c \ne d$. We need the following lemmas.

\begin{lem}\label{L:1}
A constraint language that contains $\left \{ \begin{picture}(10,10)\end{picture}_{b_0}^{a_0}\begin{picture}(10,10)\put(0,10){\line(1,-1){10}}\put(0,0){\line(1,1){10}}\put(0,10){\line(1,0){10}}\end{picture}_{b_1}^{a_1},\dots, \begin{picture}(10,10)\end{picture}_{b_{2k-1}}^{a_{2k-1}}\begin{picture}(10,10)\put(0,10){\line(1,-1){10}}\put(0,0){\line(1,1){10}}\put(0,10){\line(1,0){10}}\end{picture}_{b_{2k}}^{a_{2k}}, \begin{picture}(10,10)\end{picture}_{b_{2k}}^{a_{2k}}\begin{picture}(10,10)\put(0,10){\line(1,-1){10}}\put(0,0){\line(1,1){10}}\put(0,10){\line(1,0){10}}\end{picture}_{b_0}^{a_0}\right \}$ is NP-hard.
\end{lem}

Before proving Lemma \ref{L:1}, we need to introduce some concepts from graph theory.
All graphs are assumed to be undirected and without loops. We will be interested in
the complexity of finding independent sets of maximal size in classes of graphs.
Let a finite number of graphs $G_1 ,\dots,G_k $ be given and let $Free\left( {G_1 ,\dots,G_k } \right)$
denote the set of graphs that has no induced subgraphs isomorphic to one of $G_1 ,\dots,G_k $.

The following theorem has been proved by Alekseev\cite{alekseev}.

\begin{thm}\label{T:8}
If there is no graph among $G_1 ,\dots,G_k $ whose every connected component is a tree
with at most 3 leaves, then the maximum independent set problem is NP-hard when restricted to graphs in $Free\left( {G_1 ,\dots,G_k } \right)$.
\end{thm}

\begin{defi}\label{D:first}
The graph $G = \left( {V,E} \right)$ is said to be {\em homomorphic} to the graph $H = \left( {W,S} \right)$  if there is a mapping $f:V \to W$
such that $\forall \left( {x,y} \right) \in E\,\, \left( f( x), f( y ) \right) \in S$.
The mapping $f$ is called an {\em $H$-homomorphism}.
\end{defi}

Let $C_d $ be a cycle of length $d$.

\begin{thm}\label{T:9}
If $d \geq 3$ is odd, then the problem of finding a maximum independent set in an undirected graph homomorphic to $C_d $
 is NP-hard even if a $C_d $-homomorphism is given.
\end{thm}

\proof[Proof.]
First, we will prove NP-hardness of finding maximum independent sets in a graph homomorphic to $C_3 $, i.e. three-partite graph, following \cite{gutin0}. An instance consists of a graph and a partitioning into three independent sets.

Let $X$ be a class of graphs with degrees not greater than 3.
This class can be characterized by forbidden subgraphs ---
it is sufficient to forbid graphs with 5 vertices that has a vertex connected with 4 others.
Obviously, every such graph is connected and if it is a tree it has 4 leaves.
By Theorem \ref{T:8} we conclude that finding maximum independent sets is NP-hard in the class $X$.

From Brooks' theorem\cite{brooks}, we have that every graph in $X$, besides the complete graph on 4 vertices, is three-partite.
The required partition can be constructed in polynomial time by an algorithm of Lovasz\cite{lovasz}.
Therefore, the problem of finding maximum independent sets in a three-partite graph is NP-hard even if a partition is given.

The case when $d = 3$ can be reduced to every odd case $d > 3$.
Let a three-partite graph be given. We will define it in the following form: $G = \left( {V_1 ,V_2 ,V_3 ,E_{12} ,E_{23} ,E_{31} } \right)$, where
$E_{12}  \subseteq V_1  \times V_2 ,E_{23}  \subseteq V_2  \times V_3 ,E_{31}  \subseteq V_3  \times V_1 $.
Transform $G$ as follows: for each edge $\left( {u,v} \right) \in E_{12} $,
add vertices $x_{uv1} ,x_{uv2},\dots,x_{uv(d - 3)} $ to the graph, delete the edge $\left( {u,v} \right)$,
and add edges $\left( {u,x_{uv1} } \right),\left( {x_{uv1} ,x_{uv2} } \right)\dots,\left( {x_{uv(d - 3)} ,v} \right)$.
The obtained graph $G^d $ is, obviously, homomorphic to $C_d $.

Let $n,N$ denote the independence numbers of $G$ and $G^d $ respectively.
It is easy to see that $N \ge n + \frac{{d - 3}}{2}\left| {E_{12} } \right|$. We prove that we actually have equality there.
Note that intersection of any maximum independent set of $G^d $ and
$\left\{ {u,x_{uv1} ,x_{uv2} \dots,x_{uv(d - 3)} ,v} \right\}$ contains not less than $\frac{{d - 3}}{2}$, and not more than
$\frac{{d - 1}}{2}$ elements. In the first case($\frac{{d - 3}}{2}$), we can delete all elements $u,x_{uv1} ,x_{uv2} \dots,x_{uv(d - 3)} ,v$
from the independent set and replace them by $x_{uv1} ,x_{uv3} ,x_{uv5} ,\dots,x_{uv(d - 4)}$, while not destroying independency.
In the second case($\frac{{d - 1}}{2}$), either $u$ or $v$ are always in the independent set. Again, we delete $u,x_{uv1} ,x_{uv2} \dots,x_{uv(d - 3)} ,v$
from it. In the case where $u$ was in the independent set originally, we replace the deleted elements by $\left\{ {u,x_{uv2} ,x_{uv4} \dots,x_{uv(d - 3)} } \right\}$ and
otherwise by $\left\{ {x_{uv1} ,x_{uv3} \dots,x_{uv(d - 4)} ,v} \right\}$.
As a result, we obtain independent set of $G^d $ with the same cardinality as initially. This operation can be done with all pairs $uv\in E_{12}$.
It is easy to see that intersection of the obtained set with $V_1  \cup V_2  \cup V_3 $ is an independent set in $G$ and it
has cardinality $N - \frac{{d - 3}}{2}\left| {E_{12} } \right|$.
Consequently, $N = n + \frac{{d - 3}}{2}\left| {E_{12} } \right|$
and the constructed intersection is a maximum independent set in $G$.
The steps of construction can be carried in polynomial time. Thus, by finding a maximum independent set in $G^d $, we can easily reconstruct that of $G$. This means that the maximum independent set problem in a three-partite graph is polynomial-time reducible to the maximum independent set problem in a graph homomorphic to $C_d $(with given homomorphism).
\qed

\proof[Proof of Lemma \ref{L:1}.]

We show that finding a maximum independent set in a graph homomorphic to $C_{2k + 1}$ can be reduced to $MinHom\left( \Gamma \right)$ where
$\Gamma = \left \{
\begin{picture}(10,10)\end{picture}_{b_0}^{a_0}\begin{picture}(10,10)\put(0,10){\line(1,-1){10}}\put(0,0){\line(1,1){10}}\put(0,10){\line(1,0){10}}\end{picture}_{b_1}^{a_1}, \begin{picture}(10,10)\end{picture}_{b_1}^{a_1}\begin{picture}(10,10)\put(0,10){\line(1,-1){10}}\put(0,0){\line(1,1){10}}\put(0,10){\line(1,0){10}}\end{picture}_{b_2}^{a_2},\dots, \begin{picture}(10,10)\end{picture}_{b_{2k-1}}^{a_{2k-1}}\begin{picture}(10,10)\put(0,10){\line(1,-1){10}}\put(0,0){\line(1,1){10}}\put(0,10){\line(1,0){10}}\end{picture}_{b_{2k}}^{a_{2k}}, \begin{picture}(10,10)\end{picture}_{b_{2k}}^{a_{2k}}\begin{picture}(10,10)\put(0,10){\line(1,-1){10}}\put(0,0){\line(1,1){10}}\put(0,10){\line(1,0){10}}\end{picture}_{b_0}^{a_0}\right \}$.

Suppose the task is to find a maximum independent set in a graph homomorphic to $C_{2k + 1}$, which, for convenience, will be given
in the following form: $G = \left( {V_0 ,V_1 ,\dots,V_{2k} ,E_{i,i \oplus 1}  \subseteq V_i  \times V_{i \oplus 1} } \right)$, where $i \oplus j$ denotes $i + j(\bmod {\rm{\,\,}}2k+1)$. We consider every vertex $v\in \bigcup\limits_{i = 0}^{2k} {V_i }$ as a variable and require values of variables $(u,v)\in V_i  \times V_{i \oplus 1}$ to satisfy the constraint $\begin{picture}(10,10)\end{picture}_{b_{i}}^{a_{i}}\begin{picture}(10,10)\put(0,10){\line(1,-1){10}}\put(0,0){\line(1,1){10}}\put(0,10){\line(1,0){10}}\end{picture}_{b_{i\oplus 1}}^{a_{i\oplus 1}}$. The set of satisfying assignments is denoted by $\Phi$.
It is easy to see that
\[
\Phi = \left\{ {f|\forall v\in V_i\,\,f\left( {v } \right) \in \left\{ {a_i ,b_i } \right\},\bigcup\limits_i {\left\{ {x|x \in V_i ,f\left( x \right) = b_i } \right\}}  - {\rm{\,\,independent\,\,set\,\,in\,\,}}G} \right\}.
\]
Therefore, the task
\[
\mathop{\min}\limits_{f \in \Phi}\sum\limits_i {\sum\limits_{x \in V_i } {\left[ {f\left( x \right) \ne b_i } \right]} }
\]
is equivalent to finding a maximum independent set in the graph $G$.
I.e., it is equivalent to the $MinHom\left( H \right)$ problem with an instance consisting of the defined constraints on the variables $\bigcup\limits_{i = 0}^{2k} {V_i }$ and weights $w_{xa_i} = 1, w_{xb_i} = 0$.
Consequently, $MinHom\left( H \right)$ is NP-hard.
\qed

\begin{lem}\label{L:2}
If $\left\langle {\left( {a,b} \right),\left( {c,d} \right)} \right\rangle  \in P$,
then either $\begin{picture}(10,10)\end{picture}_b^a\begin{picture}(10,10)\put(0,10){\line(1,-1){10}}\put(0,0){\line(1,1){10}}\put(0,10){\line(1,0){10}}\end{picture}^{c}_{d} \in Inv\left ( F \right )$,
or $\begin{picture}(10,10)\end{picture}_b^a\begin{picture}(10,10)\put(0,10){\line(1,-1){10}}\put(0,0){\line(1,1){10}}\end{picture}^{c}_{d}\in Inv\left ( F \right )$.
\end{lem}

\proof[Proof.]
We begin by constructing functions $\phi _1 ,\phi _2  \in F$ such that
$\mathop  \downarrow \limits_b^a \mathop  \uparrow \limits_d^c \phi _1 ,\mathop  \uparrow \limits_b^a \mathop  \downarrow \limits_d^c \phi _2 $.
The symbol $\mathop  {\not \downarrow} \limits_\beta ^\alpha  \lambda$ means that either $\mathop  \uparrow \limits_\beta ^\alpha  \lambda$,
or $\lambda |_{\left\{ {\alpha ,\beta } \right\}} $ is a projection.

Since $\left\{ {a,b} \right\},\left\{ {c,d} \right\} \in M$, we have $\lambda _1 ,\lambda _2 ,\lambda _3 ,\lambda _4  \in F:\mathop  \downarrow \limits_b^a \lambda _1 ,\mathop  \uparrow \limits_b^a \lambda _2 ,\mathop  \downarrow \limits_d^c \lambda _3 ,\mathop  \uparrow \limits_d^c \lambda _4 $.
Moreover, by the definition of $P$, we have $\mathop {\not  \downarrow }\limits_d^c \lambda _1 ,\mathop {\not  \downarrow }\limits_b^a \lambda _3$.
By defining $\phi _1 \left( {x,y} \right) = \lambda _4 \left( {\lambda _1 \left( {x,y} \right),\lambda _1 \left( {y,x} \right)} \right),\phi _2 \left( {x,y} \right) = \lambda _2 \left( {\lambda _3 \left( {x,y} \right),\lambda _3 \left( {y,x} \right)} \right) \in F$,
we see that $\mathop  \downarrow \limits_b^a \mathop  \uparrow \limits_d^c \phi _1 ,\mathop  \uparrow \limits_b^a \mathop  \downarrow \limits_d^c \phi _2 $.

Suppose $\begin{picture}(10,10)\end{picture}_b^a\begin{picture}(10,10)\put(0,10){\line(1,-1){10}}\put(0,0){\line(1,1){10}}\put(0,10){\line(1,0){10}}\end{picture}^{c}_{d} \notin Inv\left ( F \right )$.
We prove that in this case $\begin{picture}(10,10)\end{picture}_b^a\begin{picture}(10,10)\put(0,10){\line(1,-1){10}}\put(0,0){\line(1,1){10}}\end{picture}^{c}_{d}\in Inv\left ( F \right )$.
Since the predicate $\begin{picture}(10,10)\end{picture}_b^a\begin{picture}(10,10)\put(0,10){\line(1,-1){10}}\put(0,0){\line(1,1){10}}\put(0,10){\line(1,0){10}}\end{picture}^{c}_{d}$
consists of three pairs, it is not preserved by some function of arity two or three. Let us consider these two cases:

I. A function $\phi \in F$ of arity two does not preserve $\begin{picture}(10,10)\end{picture}_b^a\begin{picture}(10,10)\put(0,10){\line(1,-1){10}}\put(0,0){\line(1,1){10}}\put(0,10){\line(1,0){10}}\end{picture}^{c}_{d}$ if (for some appropriate permutation of variables):
\[
\begin{array}{*{20}c}
   {\phi \left( {a,b} \right) = b}  \\
   {\phi \left( {d,c} \right) = d}  \\
\end{array}.
\]
Then $\mathop  \downarrow \limits_b^a \mathop  \downarrow \limits_d^c \phi \left( {\phi _2 \left( {x,y} \right),\phi _1 \left( {x,y} \right)} \right)$
which contradicts that $\left\langle {\left( {a,b} \right),\left( {c,d} \right)} \right\rangle  \in P$.

II. A function $\phi \in F$ of arity three does not preserve $\begin{picture}(10,10)\end{picture}_b^a\begin{picture}(10,10)\put(0,10){\line(1,-1){10}}\put(0,0){\line(1,1){10}}\put(0,10){\line(1,0){10}}\end{picture}^{c}_{d}$  if (for some appropriate permutation of variables):
\[
\begin{array}{*{20}c}
   {\phi \left( {a,a,b} \right) = b}  \\
   {\phi \left( {d,c,c} \right) = d}  \\
\end{array}.
\]
Then, $\left\langle {\left( {b,a} \right),\left( {d,c} \right)} \right\rangle  \in P$,
since, otherwise, we can find $\phi _3\in F :\mathop  \uparrow \limits_b^a \mathop  \uparrow \limits_d^c \phi _3 $ and construct the following term
$\mathop  \downarrow \limits_b^a \mathop  \downarrow \limits_d^c \phi \left( {\phi _2 \left( {x,y} \right),\phi _3 \left( {x,y} \right),\phi _1 \left( {x,y} \right)} \right)$. This contradicts that $\left\langle {\left( {a,b} \right),\left( {c,d} \right)} \right\rangle  \in P$.
Suppose instead that $\begin{picture}(10,10)\end{picture}_b^a\begin{picture}(10,10)\put(0,10){\line(1,-1){10}}\put(0,0){\line(1,1){10}}\end{picture}^{c}_{d}\notin Inv\left ( F \right )$, i.e., there is a function $f\in F$ of arity two that does not preserve $\begin{picture}(10,10)\end{picture}_b^a\begin{picture}(10,10)\put(0,10){\line(1,-1){10}}\put(0,0){\line(1,1){10}}\end{picture}^{c}_{d}$. If $f$ does not preserve
$\begin{picture}(10,10)\end{picture}_b^a\begin{picture}(10,10)\put(0,10){\line(1,-1){10}}\put(0,0){\line(1,1){10}}\end{picture}^{c}_{d}$,
then it does not preserve either
$\begin{picture}(10,10)\end{picture}_b^a\begin{picture}(10,10)\put(0,10){\line(1,-1){10}}\put(0,0){\line(1,1){10}}\put(0,10){\line(1,0){10}}\end{picture}^{c}_{d}$, or
$\begin{picture}(10,10)\end{picture}_a^b\begin{picture}(10,10)\put(0,10){\line(1,-1){10}}\put(0,0){\line(1,1){10}}\put(0,10){\line(1,0){10}}\end{picture}^{d}_{c}$.
Since $\left\langle {\left( {a,b} \right),\left( {c,d} \right)} \right\rangle, \left\langle {\left( {b,a} \right),\left( {d,c} \right)} \right\rangle  \in P$,
we get a contradiction in both cases via the same argument as in case I.
\qed

\proof[Proof of NP-hard case of Theorem \ref{T:7}.] For binary predicates $\alpha, \beta$, let $\alpha \circ \beta = \left\{(x,y)| \exists z: \alpha(x,z) \wedge \beta(z,y)\right\}$. Obviously, if $\alpha, \beta \in Inv\left( {F } \right)$, then $\alpha \circ \beta \in Inv\left( {F } \right)$, too.

Since $T_F = \left( {M^o ,P} \right)$
is not bipartite, we can find a shortest odd cycle in it, i.e. a sequence
$\left( {a_0 ,b_0 } \right),\left( {a_1 ,b_1 } \right),\dots,\left( {a_{2k} ,b_{2k} } \right) \in M^o ,k \ge 1$, such that $\left\langle {\left( {a_i ,b_i } \right),\left( {a_{i \oplus 1} ,b_{i \oplus 1} } \right)} \right\rangle  \in P$. Here, $i \oplus j$ denotes $i + j(\bmod {\rm{\,\,}}2k + 1)$.

By Lemma \ref{L:2}, there is a cyclic sequence
$\rho _{0,1 },\rho _{1,2 },\dots,\rho _{2k,0}\in Inv\left( F \right)$
such that $\rho _{i,i \oplus 1}$ is either equal to
$\begin{picture}(10,10)\end{picture}_{b_i}^{a_i}\begin{picture}(10,10)\put(0,10){\line(1,-1){10}}\put(0,0){\line(1,1){10}}\put(0,10){\line(1,0){10}}\end{picture}_{b_{i\oplus 1}}^{a_{i\oplus 1}} $
or equal to $\begin{picture}(10,10)\end{picture}_{b_i}^{a_i}\begin{picture}(10,10)\put(0,10){\line(1,-1){10}}\put(0,0){\line(1,1){10}}\end{picture}_{b_{i\oplus 1}}^{a_{i\oplus 1}} $.
Note that all predicates cannot be of the second type: otherwise, we have $
\rho _{0,1} \circ \rho _{1,2} \circ \dots \circ \rho _{2k,0} = \begin{picture}(10,10)\end{picture}_{b_0}^{a_0}\begin{picture}(10,10)\put(0,10){\line(1,-1){10}}\put(0,0){\line(1,1){10}}\end{picture}_{b_0}^{a_0}$
which contradicts that $\left\{ {a_0 ,b_0 } \right\} \in M$.

If the sequence contains a fragment
$\rho {}_{i,i \oplus 1 }  = \begin{picture}(10,10)\end{picture}_{b_i}^{a_i}\begin{picture}(10,10)\put(0,10){\line(1,-1){10}}\put(0,0){\line(1,1){10}}\end{picture}_{b_{i\oplus 1}}^{a_{i\oplus 1}}$,
$\rho {}_{i\oplus 1,i\oplus 2}  = \begin{picture}(10,10)\end{picture}_{b_{i\oplus 1}}^{a_{i\oplus 1}}\begin{picture}(10,10)\put(0,10){\line(1,-1){10}}\put(0,0){\line(1,1){10}}\end{picture}_{b_{i\oplus 2}}^{a_{i\oplus 2}}$,
$\rho {}_{i\oplus 2,i \oplus 3}  = \begin{picture}(10,10)\end{picture}_{b_{i\oplus 2}}^{a_{i\oplus 2}} \begin{picture}(10,10)\put(0,10){\line(1,-1){10}}\put(0,0){\line(1,1){10}}\put(0,10){\line(1,0){10}}\end{picture}_{b_{i\oplus 3}}^{a_{i\oplus 3}}$,
then these predicates can be replaced by:
\[
\rho {}_{i,i \oplus 3 }  \buildrel \Delta \over = \rho {}_{i,i \oplus 1}  \circ \rho {}_{i \oplus 1,i \oplus 2 }  \circ \rho {}_{i \oplus 2, i \oplus 3 }  =
\begin{picture}(10,10)\end{picture}_{b_i}^{a_i}\begin{picture}(10,10)\put(0,10){\line(1,-1){10}}\put(0,0){\line(1,1){10}}\end{picture}_{b_{i\oplus 1}}^{a_{i\oplus 1}} \circ
\begin{picture}(10,10)\end{picture}_{b_{i\oplus 1}}^{a_{i\oplus 1}}\begin{picture}(10,10)\put(0,10){\line(1,-1){10}}\put(0,0){\line(1,1){10}}\end{picture}_{b_{i\oplus 2}}^{a_{i\oplus 2}} \circ
\begin{picture}(10,10)\end{picture}_{b_{i\oplus 2}}^{a_{i\oplus 2}} \begin{picture}(10,10)\put(0,10){\line(1,-1){10}}\put(0,0){\line(1,1){10}}\put(0,10){\line(1,0){10}}\end{picture}_{b_{i\oplus 3}}^{a_{i\oplus 3}} =
\begin{picture}(10,10)\end{picture}_{b_i}^{a_i} \begin{picture}(10,10)\put(0,10){\line(1,-1){10}}\put(0,0){\line(1,1){10}}\put(0,10){\line(1,0){10}}\end{picture}_{b_{i\oplus 3}}^{a_{i\oplus 3}}
\]
Let us replace $\rho {}_{i,i \oplus 1 }$, $\rho {}_{i \oplus 1,i \oplus 2 }$, $\rho {}_{i \oplus 2, i \oplus 3 }$
by $\rho {}_{i,i \oplus 3 }$ in the sequence
$\rho _{0,1 },\rho _{1,2 },\dots,\rho _{2k,0} $.
We have $\left\langle {\left( {a_{i} ,b_{i} } \right),\left( {a_{i \oplus 3} ,b_{i \oplus 3} } \right)} \right\rangle  \in P$,
since otherwise the predicate $\rho {}_{i, i \oplus 3}$ is not preserved.
Hence, we can delete two vertices in the cycle
$\left( {a_0 ,b_0 } \right),\left( {a_1 ,b_1 } \right),\dots,\left( {a_{2k} ,b_{2k} } \right) \in M^o $.
This contradicts that this sequence is the shortest among odd sequences.
Therefore, such a fragment does not exist.

If the sequence contains a fragment
$\rho {}_{i,i \oplus 1}  = \begin{picture}(10,10)\end{picture}_{b_{i}}^{a_{i}} \begin{picture}(10,10)\put(0,10){\line(1,-1){10}}\put(0,0){\line(1,1){10}}\put(0,10){\line(1,0){10}}\end{picture}_{b_{i\oplus 1}}^{a_{i\oplus 1}}$, $\rho {}_{i\oplus 1,i \oplus 2} = \begin{picture}(10,10)\end{picture}_{b_{i\oplus 1}}^{a_{i\oplus 1}}\begin{picture}(10,10)\put(0,10){\line(1,-1){10}}\put(0,0){\line(1,1){10}}\end{picture}_{b_{i\oplus 2}}^{a_{i\oplus 2}}$, $\rho {}_{i\oplus 2, i \oplus 3}  = \begin{picture}(10,10)\end{picture}_{b_{i\oplus 2}}^{a_{i\oplus 2}} \begin{picture}(10,10)\put(0,10){\line(1,-1){10}}\put(0,0){\line(1,1){10}}\put(0,10){\line(1,0){10}}\end{picture}_{b_{i\oplus 3}}^{a_{i\oplus 3}}$,
then these predicates can be replaced by:
\[
\rho {}_{i, i \oplus 3 } \buildrel \Delta \over = \rho {}_{i,i \oplus 1}  \circ \rho {}_{i \oplus 1,i \oplus 2 }  \circ \rho {}_{i \oplus 2, i \oplus 3 }  =
\begin{picture}(10,10)\end{picture}_{b_{i}}^{a_{i}} \begin{picture}(10,10)\put(0,10){\line(1,-1){10}}\put(0,0){\line(1,1){10}}\put(0,10){\line(1,0){10}}\end{picture}_{b_{i\oplus 1}}^{a_{i\oplus 1}} \circ
\begin{picture}(10,10)\end{picture}_{b_{i\oplus 1}}^{a_{i\oplus 1}}\begin{picture}(10,10)\put(0,10){\line(1,-1){10}}\put(0,0){\line(1,1){10}}\end{picture}_{b_{i\oplus 2}}^{a_{i\oplus 2}} \circ
\begin{picture}(10,10)\end{picture}_{b_{i\oplus 2}}^{a_{i\oplus 2}} \begin{picture}(10,10)\put(0,10){\line(1,-1){10}}\put(0,0){\line(1,1){10}}\put(0,10){\line(1,0){10}}\end{picture}_{b_{i\oplus 3}}^{a_{i\oplus 3}} =
\begin{picture}(10,10)\end{picture}_{b_i}^{a_i} \begin{picture}(10,10)\put(0,10){\line(1,-1){10}}\put(0,0){\line(1,1){10}}\put(0,10){\line(1,0){10}}\end{picture}_{b_{i\oplus 3}}^{a_{i\oplus 3}}
\]
As in the previous case, we obtain a contradiction. Consequently, we have an odd sequence
$\begin{picture}(10,10)\end{picture}_{b_0}^{a_0}\begin{picture}(10,10)\put(0,10){\line(1,-1){10}}\put(0,0){\line(1,1){10}}\put(0,10){\line(1,0){10}}\end{picture}_{b_1}^{a_1}, \begin{picture}(10,10)\end{picture}_{b_1}^{a_1}\begin{picture}(10,10)\put(0,10){\line(1,-1){10}}\put(0,0){\line(1,1){10}}\put(0,10){\line(1,0){10}}\end{picture}_{b_2}^{a_2},\dots, \begin{picture}(10,10)\end{picture}_{b_{2k-1}}^{a_{2k-1}}\begin{picture}(10,10)\put(0,10){\line(1,-1){10}}\put(0,0){\line(1,1){10}}\put(0,10){\line(1,0){10}}\end{picture}_{b_{2k}}^{a_{2k}}, \begin{picture}(10,10)\end{picture}_{b_{2k}}^{a_{2k}}\begin{picture}(10,10)\put(0,10){\line(1,-1){10}}\put(0,0){\line(1,1){10}}\put(0,10){\line(1,0){10}}\end{picture}_{b_0}^{a_0}  \in Inv\left( F \right)
$. By Lemma \ref{L:1}, this class of predicates is NP-hard.
\qed

\section{Existence of the majority operation}
The necessary local conditions tell that every two-element subalgebra of a tractable algebra contains certain operations. The simplest algebras over a domain $A$ that satisfy these conditions are the following: $F_1=\left\{\phi, \psi\right\}$ where $\phi, \psi$ are conservative commutative operations such that $\phi(a,b)\ne\psi(a,b)$ for every $a\ne b\in A$, and $F_2=\left\{m\right\}$ where $m$ is a conservative arithmetical operation, i.e. $m\left( {x,x,y} \right) = m \left( {y,x,x} \right) = m \left( {y,x,y} \right) = y$.
This leads us to the following definitions.

\begin{defi}\label{D:first}
Suppose a set of operations $H$ over $D$ is conservative and $B\subseteq \left\{ {\left\{ {x,y} \right\}|x,y \in D ,x \ne y} \right\}$. A pair of binary operations $\phi, \psi \in H$ is called a {\em tournament pair} on $B$,
if $\forall \left\{ {x,y} \right\} \in B {\rm{\,\,}}\phi \left( {x,y} \right) = \phi \left( {y,x} \right),\psi \left( {x,y} \right) = \psi \left( {y,x} \right), {\phi \left( {x,y} \right)\ne\psi \left( {x,y} \right)}$
and for arbitrary $\left\{ {x,y} \right\} \in \overline{B} $, $\phi \left( {x,y} \right) = x,\psi \left( {x,y} \right) = x$.
An operation  $m \in H$ is called {\em arithmetical} on $B$, if
$\forall \left\{ {x,y} \right\} \in B {\rm{\,\,}}m \left( {x,x,y} \right) = m \left( {y,x,x} \right) = m \left( {y,x,y} \right) = y$.
\end{defi}

\begin{defi}\label{D:first}
An operation $\mu :A^3  \to A$, satisfying the equality
\[
\mu \left( {x,y,y} \right) = \mu \left( {y,x,y} \right) = \mu \left( {y,y,x} \right) = y
\]
is called a majority operation.
\end{defi}

\begin{thm}\label{T:11}
If $F$ satisfies the necessary local conditions and $T_F = \left( {M^o ,P} \right)$ is bipartite, then $F$ contains a tournament pair on $M$.
\end{thm}

\proof[Proof.]
Let $M_1, M_2$ denote a partitioning of the bipartite graph $T_F = \left( {M^o ,P} \right)$.
Then, for every $\left( {a,b} \right),\left( {c,d} \right) \in M_1  $,
there is a function $\phi\in F :\mathop  \downarrow \limits_b^a \mathop  \downarrow \limits_d^c \phi $.
Let us prove by induction that for every
$\left( {a_1 ,b_1 } \right),\left( {a_2 ,b_2 } \right),\dots,\left( {a_n ,b_n } \right) \in M_1  $,
there is a $\phi :\mathop  \downarrow \limits_{b_1 }^{a_1 } \mathop  \downarrow \limits_{b_2 }^{a_2 } \dots\mathop  \downarrow \limits_{b_n }^{a_n } \phi $.

The base of induction $n = 2$ is obvious.
Let $\left( {a_1 ,b_1 } \right),\left( {a_2 ,b_2 } \right),\dots,\left( {a_{n + 1} ,b_{n + 1} } \right) \in M_1  $ be given.
By the induction hypothesis, there are  $
\phi _1 ,\phi _2 ,\phi _3\in F :\mathop  \downarrow \limits_{b_2 }^{a_2 } \dots\mathop  \downarrow \limits_{b_n }^{a_n } \mathop  \downarrow \limits_{b_{n + 1} }^{a_{n + 1} } \phi _1 ,\mathop  \downarrow \limits_{b_1 }^{a_1 } \mathop  \downarrow \limits_{b_3 }^{a_3 } \dots\mathop  \downarrow \limits_{b_n }^{a_n } \mathop  \downarrow \limits_{b_{n + 1} }^{a_{n + 1} } \phi _2 ,\mathop  \downarrow \limits_{b_1 }^{a_1 } \mathop  \downarrow \limits_{b_2 }^{a_2 } \dots\mathop  \downarrow \limits_{b_n }^{a_n } \phi _3$.
Then, it is easy to see that
$\mathop  \downarrow \limits_{b_1 }^{a_1 } \dots\mathop  \downarrow \limits_{b_n }^{a_n } \mathop  \downarrow \limits_{b_{n + 1} }^{a_{n + 1} } \phi _3 \left( {\phi _1 \left( {x,y} \right),\phi _2 \left( {x,y} \right)} \right)$ which completes the induction proof.

The analogous statement can be proved for $M_2$. Moreover, $M_2   = \left\{ {\left( {x,y} \right)|\left( {y,x} \right) \in M_1  } \right\}$.
So it follows from the proof that there are binary operations
$\phi ',\psi '\in F$, such that
$\forall \left( {x,y} \right) \in M_1 {\rm{: }}\mathop  \downarrow \limits_y^x \phi '$ and $\forall \left( {x,y} \right) \in M_2 {\rm{: }}\mathop  \downarrow \limits_y^x \psi '$.
Thus, the operations $\phi \left( {x,y} \right) = \phi '\left( {x,\phi '\left( {y,x} \right)} \right)$ and $\psi \left( {x,y} \right) = \psi '\left( {x,\psi '\left( {y,x} \right)} \right)$
satisfy the conditions of theorem.
\qed

The proof of the following theorem uses ideas from \cite{bulatov}.

\begin{thm}\label{T:12}
If $F$ satisfies the necessary local conditions and
$\overline M  \ne \emptyset $, then $F$ contains an arithmetical operation on $\overline M$.
\end{thm}

\proof[Proof.]
Note first that for every $B \in \overline M $, $F|_B $ cannot contain any commutative binary function.
To see this, assume that $B = \{0, 1\}$ and note that $F|_B $ contains $S_{01} $ and either
conjunction or disjunction. From Post's results \cite{post}, we see that $F|_B $ contains all boolean functions preserving 0 and 1, i.e.,
contains both conjunction and disjunction.
This contradicts that $B \notin M$.
Therefore, every binary function in $F|_B $ is a projection.

For $B \in \overline M$, let $m^B$ be an arithmetical function on $B$; existence of this function follows from the necessary local conditions.
Assume now that $\overline M  = \left\{ {\left\{ {x_1 ,y_1 } \right\},\dots,\left\{ {x_s ,y_s } \right\}} \right\}$.
We prove by induction that for every $r \le s$, $F$ contains a function $m_r :A^3  \to A$ that is arithmetical on
$\left\{ \left\{ {x_i ,y_i } \right\} | 1\leq i \leq r\right\}$.

When $r = 1$, $m_1 \left( {x,y,z} \right) = m^{\left\{ {x_1 ,y_1 } \right\}} \left( {x,y,z} \right)$ and the statement is
obviously true.
Suppose it is true for $r \le k < s$ and that we have the function $m_k :A^3  \to A$.
Let us prove the statement for $r = k + 1$. If $m_k$ is arithmetical on $\left\{\left\{ {x_{k + 1} ,y_{k + 1} } \right\}\right\}$, then we define $m_{k + 1} = m_k $
and the statement is proved. Otherwise, one of the following three statements is true
\[
\exists x,y \in \left\{ {x_{k + 1} ,y_{k + 1} } \right\}{\rm{ }}\left[ {m_k \left( {x,x,y} \right) \ne y} \right],
\]
\[
\exists x,y \in \left\{ {x_{k + 1} ,y_{k + 1} } \right\}{\rm{ }}\left[ {m_k \left( {y,x,x} \right) \ne y} \right],
\]
\[
\exists x,y \in \left\{ {x_{k + 1} ,y_{k + 1} } \right\}{\rm{ }}\left[ {m_k \left( {y,x,y} \right) \ne y} \right].
\]

Suppose the first case holds (the proof for other cases is analogous), i.e.
$m_k |_{\left\{ {x_{k + 1} ,y_{k + 1} } \right\}} \left( {x,x,y} \right)$ is the $x$-projection.
It is easy to see that the function $m_{k + 1} \left( {x,y,z} \right) = m_k \left( {m^{\left\{ {x_{k + 1} ,y_{k + 1} } \right\}} \left( {x,y,z} \right),m^{\left\{ {x_{k + 1} ,y_{k + 1} } \right\}} \left( {x,y,z} \right),m_k \left( {x,y,z} \right)} \right)$
is arithmetical on $\left\{ \left\{ {x_i ,y_i } \right\} | 1\leq i \leq k+1\right\}$.

Induction completed and it is clear that $m_s \left( {x,y,z} \right)$ satisfies the condition of theorem.
\qed

\begin{thm}\label{T:13}
If $F$ satisfies the necessary local conditions and $T_F = \left( {M^o ,P} \right)$ is bipartite, then
$F$ contains a majority operation $\mu $.
\end{thm}

\proof[Proof.]
If $\overline M  \ne \emptyset $, then by Theorem \ref{T:12}, $F$ contains a function $m:A^3  \to A$ that is arithmetical on $\overline M$.
Then the function $\mu ^1 \left( {x,y,z} \right) = m\left( {x,m\left( {x,y,z} \right),z} \right)$ satisfies the conditions
$\forall \left\{ {x,y} \right\} \in \overline M {\rm{\,\,}}\mu ^1 \left( {x,y,y} \right) = \mu ^1 \left( {y,x,y} \right) = \mu ^1 \left( {y,y,x} \right) = y$.
It is clear that, in the case where $M = \emptyset $, we can take $\mu ^1 $ as majority $\mu $.

If $M \ne \emptyset $, then by Theorem \ref{T:11}, there is a tournament pair $\phi ,\psi :A^2  \to A$ on $M$.
Then, the function $\mu ^2 \left( {x,y,z} \right) = \phi \left( {\phi \left( {\psi \left( {x,y} \right),\psi \left( {y,z} \right)} \right),\psi \left( {x,z} \right)} \right)$ satisfies conditions
$\forall \left\{ {x,y} \right\} \in M{\rm{\,\,}}\mu ^2 \left( {x,y,y} \right) = \mu ^2 \left( {y,x,y} \right) = \mu ^2 \left( {y,y,x} \right) = y$,
and $\forall \left\{ {x,y,z} \right\} \in \overline M {\rm{\,\,}}\mu ^2 \left( {x,y,z} \right) = x$.
If $\overline M  = \emptyset $, then we can take $\mu ^2 $ as the majority $\mu $.

Finally, if $M,\overline M  \ne \emptyset $,
then $\mu \left( {x,y,z} \right) = \mu ^1 \left( {\mu ^2 \left( {x,y,z} \right),\mu ^2 \left( {y,z,x} \right),\mu ^2 \left( {z,x,y} \right)} \right)$.
\qed

\section{Consistency and microstructure graphs}
Every predicate in $Inv\left( F \right)$, when $F$ contains a majority operation, is equal to the join
of its binary projections \cite{baker}.
To prove Theorem \ref{T:7}, it is consequently sufficient to prove polynomial-time solvability of
$MinHom\left( {\Gamma} \right)$ where $\Gamma = \left\{ \rho | \rho  \subseteq A^2 ,\rho  \in Inv\left( F \right)\right\} $, i.e. the $MinHom$ problem
restricted to binary constraint languages.

\begin{defi}\label{D:first}
Suppose we are given a constraint language $\Gamma$ over $A$. Denote by $2-MinHom\left( \Gamma \right)$ the following minimization problem:

\noindent {\bf Instance:} A finite set of variables $X = \left\{ x_1 ,\dots,x_n \right\}$, a constraints pair $\left( {U,B } \right)$ where $U = \langle {\rho _i } \rangle_{1 \le i \le n}$, $B = \langle {\rho _{kl} } \rangle_{1 \le k \ne l \le n} $, $\rho _i, \rho _{kl}\in \Gamma$, and weights $w_{ia} ,1 \le i \le n ,a \in A$.

\noindent {\bf Solution:} Assignment $f:\left\{ {{x_1},\dots,{x_n}} \right\} \rightarrow A$, such that $\forall i{\rm{\,\,}}f\left( x_i \right)  \in \rho _i$ and $\forall k \ne l{\rm{\,\,}}\left( {f\left( x_k \right),f\left( x_l \right) } \right) \in \rho _{kl}$.

\noindent {\bf Measure:} $\sum\limits_{i = 1}^n {w_{if\left( x_{i} \right)}}$.
\end{defi}

We suppose everywhere that $\rho _{kl}  = \rho _{lk}^t$ (where $\rho ^t  = \left\{ {\left( {b,a} \right)|\left( {a,b} \right) \in \rho } \right\}$).
If $\rho _{kl}  \ne \rho _{lk}^t$, then we can always define $\forall k \ne l{\rm{\,\,}}\rho _{kl} : = \rho _{kl}  \cap \rho _{lk}^t$,
which does not change the set $\left\{ {\left({ a ,b } \right)|\left( {a ,b } \right) \in \rho _{kl} ,\left( {b ,a } \right) \in \rho _{lk} } \right\}$. For a binary predicate $\rho$, define projections $\Pr _1 \rho = \left\{a|(a,b)\in \rho\right\}$ and $\Pr _2 \rho = \left\{b|(a,b)\in \rho\right\}$.

\begin{defi}\label{D:first}
An instance of $2-MinHom\left( \Gamma \right)$ with constraints pair $U = \langle {\rho _i } \rangle_{1 \le i \le n}$, $B = \langle {\rho _{kl} } \rangle_{1 \le k \ne l \le n} $
is called {\em arc-consistent} if $\forall i \ne j:{\rm{ }}\Pr _1 \rho _{ij}  = \rho _i ,\Pr _2 \rho _{ij}  = \rho _j$ and
is called {\em path-consistent} if for each different $i,j,k:{\rm{ }}\rho _{ik}  \subseteq \rho _{ij}  \circ \rho _{jk} $.
\end{defi}

Obviously, by applying operations
$\rho _i : = \rho _i  \cap \Pr _1 \rho _{ij}$, $\rho _j : = \rho _j  \cap \Pr _2 \rho _{ij}$,
$\rho _{ij} : = \rho _{ij}  \cap \left( {\rho _i  \times A} \right)$, $\rho _{ij} : = \rho _{ij}  \cap \left( {A \times \rho _j } \right)$,
$\rho _{ik} : = \rho _{ik}  \cap \left( {\rho _{ij}  \circ \rho _{jk} } \right)$,
we can always make an instance arc-consistent and path-consistent in polynomial time. It is clear that under this transformations the set of feasible solutions does not change.

\begin{defi}\label{D:first}
The {\em microstructure graph} \cite{jegou} of an instance of $2-MinHom\left( \Gamma \right)$ with constraints pair
$U = \langle {\rho _i } \rangle_{1 \le i \le n}$, $B = \langle {\rho _{kl} } \rangle_{1 \le k \ne l \le n} $
is the graph $M_{U,B} = \left( {V,E} \right)$, where $V = \left\{ {\left( {i,a} \right)|1\leq i\leq n ,a \in \rho _i } \right\}$
and $E =\left\{ {\left\langle {\left( {i,a} \right),\left( {j,b} \right)} \right\rangle |i \ne j,\left( {a,b} \right) \in \rho _{ij} } \right\}$.
\end{defi}

\begin{thm}\label{T:VVG}
Let $I = \left( X, U, B, w\right)$ be a satisfiable instance of $2-MinHom\left( \Gamma \right)$. Then there is a one-to-one correspondence between
maximal-size cliques of $M_{U,B}$ and satisfying assignments of $I$.
\end{thm}

\proof[Proof.] The microstructure graph of an instance with constraints pair
$U = \langle {\rho _i } \rangle_{1 \le i \le n}$, $B = \langle {\rho _{kl} } \rangle_{1 \le k \ne l \le n} $ is, obviously, $n$-partite,
since $V = \bigcup\limits_{i = 1}^n {\left\{ i \right\} \times \rho _i} $
and pairs $\left( {i,a} \right),\left( {i,b} \right),a \ne b$ are not connected.
Therefore, the cardinality of a maximal clique of $M_{U,B} = \left( {V,E} \right)$ is not greater than $n$.

If the cardinality of a maximal clique $S \subseteq V$ is $n$, then, for every $i$,
$\left| {S \cap \left(\left\{ i \right\} \times \rho _i\right) }  \right| = 1$.
Then, denoting the only element of $S \cap \left(\left\{ i \right\} \times \rho _i\right)$ by $v_i $,
we see that the assignment $f\left(x_i\right)  = v_i $ satisfies all constraints.
The opposite is also true, i.e., if the constraints
$\langle {\rho _i } \rangle_{1 \le i \le n} ,\langle {\rho _{kl} } \rangle_{1 \le k \ne l \le n} $
can be satisfied by some assignment $f$, then $\left\{ {\left( {i,f\left(x_i\right) } \right)|1\leq i\leq n } \right\}$ is
a clique of cardinality $n$.
\qed

%\begin{rem}\label{R:1}
%An instance is satisfiable if and only if the chromatic number of the microstructure graph $M_{U,B}$ is equal to a maximal clique number. As %well-known, equality of chromatic number and maximal clique number for any induced subgraph of a graph is the definition of perfectness. The %microstructure graph of an instance, is in general not perfect, but this case, as we will see, is crucial for the tractability of $2 - MinHom$, and %therefore also for the tractability of $MinHom$.
%\end{rem}

Hence, $2-MinHom\left( \Gamma \right)$ can be reduced to finding a maximal-size clique $S \subseteq V$
of a microstructure graph that minimizes the following value:
\[
\sum\limits_{\left( {i,a} \right) \in S} {w_{ia} }.
\]

\begin{defi}\label{D:first}
Let $MMClique$ (Minimal weight among maximal-size cliques) denote the following minimization problem:

\noindent {\bf Instance:} A graph $G = \left( {V,E} \right)$ and weights $w_i  \in {\mathbb N},i \in V$.

\noindent {\bf Solution:} A maximal-size clique $K \subseteq V$ of $G$.

\noindent {\bf Measure:} $\sum\limits_{v\in K} {w_{v}}$.
\end{defi}

The following theorem connects perfect microstructure graphs and the complexity of $MinHom$.

\begin{thm}\label{T:perfect}
Suppose we are given a class of conservative functions $F$ containing a majority operation. If the microstructure graph is perfect
for arbitrary arc-consistent and path-consistent instances of $2 - MinHom\left( {Inv\left( {F} \right)} \right)$, then $F$ is tractable.
\end{thm}

\proof[Proof.] Recall that a graph $G = \left( {V,E} \right)$
is called perfect if for every induced subgraph the chromatic number is equal to the
clique number.

For a graph $G = \left( {V,E} \right)$, the following polytope is called the
{\em fractional stable set polytope}:

$\left\{ {\begin{array}{*{20}c}
   {\sum\limits_{v \in K} {x_v }  \le 1,{\rm{\,\,where\,\,}}K{\rm{\,\,is\,\,a\,\,clique\,\,in\,\,}G}}  \\
   {x_v  \ge 0,v \in V}  \\
\end{array}} \right.$

By a well-known theorem of Lovasz\cite{grotshel}, a graph $G = \left( {V,E} \right)$ is perfect if and only if
its fractional stable set polytope equals the convex hull of the characteristic vectors of independent sets in $G$.
By the vertex packing problem we mean the weighted version of maximum independent set.
It is easy to see that vertex packing in perfect graphs is equivalent to optimizing a linear function over the
fractional stable set polytope. There is a polynomial algorithm for solving the vertex packing in perfect graphs\cite{grotshel2}.
Using well-known results\cite{grotshel,hachiyan} about polynomial equivalence between the separation and
optimization of linear function on polytopes we obtain that
there is a polynomial algorithm that takes a perfect graph $G = \left( {V,E} \right)$, a rational
vector $a_v ,v \in V$ as input, and checks whether the vector is in the fractional stable set polytope
of $G$ or not. If not, it finds a hyperplane (given by rational vectors) that separates $a_v ,v \in V$ from the polytope.

Therefore, there exists a polynomial separation algorithm for the fractional stable set polytope of a perfect graph with
addition of the following equality:
$\sum\limits_{v \in V} {x_v }  = \alpha \left( G \right)$ where $\alpha \left( G \right)$ is independence number of the given graph $G$.
That is, we have a polynomial algorithm for the following task:

$\left\{ {\begin{array}{*{20}c}
   {\sum\limits_{v \in K} {x_v }  \le 1,{\rm{\,\,where\,\,}}K{\rm{\,\,is\,\,a\,\,clique\,\,in\,\,}G}}  \\
   {x_v  \ge 0,v \in V}  \\
   {\sum\limits_{v \in V} {x_v }  = \alpha \left( G \right)}  \\
   {\sum\limits_{v \in V} {w_v x_v }  \to \min}  \\
\end{array}} \right.$

It is easy to see that this task coincides with MMClique for the complement of $G$. Since the complement of a
perfect graph is perfect, MMClique for perfect graphs is polynomial-time solvable, too.
\qed

\begin{defi}\label{D:first}
A cycle $C_{2k + 1}$, $k \ge 2$, is called {\em an odd hole} and its complement graph {\em an odd antihole}.
\end{defi}

In Section 8 we will use the following conjecture of Berge, which was proved in \cite{chudnovsky}.

\begin{thm}\label{T:14}
A graph is perfect if and only if it does not contain an induced subgraph isomorphic to an odd hole or antihole.
\end{thm}

We say that a graph is {\em of type} $S_{2k + 1} ,k \ge 2$ if it is isomorphic to the graph with vertex set
$\left\{ {0,1,\dots,2k} \right\}$, where vertices $i\left( {\bmod {\rm{\,\,}}2k + 1} \right)$, $i + 1\left( {\bmod {\rm{\,\,}}2k + 1} \right)$
are not connected and vertices $i\left( {\bmod {\rm{\,\,}}2k + 1} \right)$, $i + 2\left( {\bmod {\rm{\,\,}}2k + 1} \right)$ are
connected. Other pairs can be connected arbitrarily.
Obviously, every odd hole or antihole is of one of types $S_{2k + 1} ,k \ge 2$.

\section{Arithmetical deadlocks}
The key idea for the proof of the polynomial case of Theorem \ref{T:7} is to show that path- and arc-consistent instances of $2 - MinHom\left( {Inv\left( {F} \right)} \right)$ have a perfect microstructure graph. We will prove this by showing that the microstructure graph forbids certain types of subgraphs. The exact formulation of the result can be found below in Theorem \ref{T:16}. This theorem uses the nonexistence of structures called {\em arithmetical deadlocks} which are introduced in this section.

\begin{defi}\label{D:first}
Suppose $H$ is a conservative set of functions over $D$, $m \in H$ is an arithmetical operation on $B\subseteq \left\{ {\left\{ {x,y} \right\}|x,y \in D ,x \ne y} \right\}$ and a pair
$\phi, \psi \in H$ is a tournament pair on $\overline{B}$. An instance of $2-MinHom\left( Inv\left( H \right) \right)$ with constraints pair $U = \langle {\rho _i } \rangle_{1 \le i \le n}$, $B = \langle {\rho _{kl} } \rangle_{1 \le k \ne l \le n} $ is called {\em an odd arithmetical deadlock} if there is a subset
$\left\{ {i_0 ,\dots,i_{k - 1} } \right\} \subseteq \left\{ {1 ,\dots,n } \right\},k \ge 3$ of odd cardinality and
$\left\{ {x_0 ,y_0 } \right\},\dots,\left\{ {x_{k - 1} ,y_{k - 1} } \right\} \in B$, such that for
$0\leq s \leq k - 1 $:  $\rho _{i{}_s,i{}_{s \oplus 1}}  \cap \left\{ {x_s ,y_s } \right\} \times \left\{ {x_{s \oplus 1} ,y_{s \oplus 1} } \right\} = \begin{picture}(10,10)\end{picture}_{ y_s }^{ x_s }\begin{picture}(10,10)\put(0,10){\line(1,-1){10}}\put(0,0){\line(1,1){10}}\end{picture}^{ x_{s \oplus 1}}_{ y_{s \oplus 1}}$, where $i \oplus j$ denotes $i + j(\bmod {\rm{\,\,}}k)$.
The subset $\left\{ {i_0 ,\dots,i_{k - 1} } \right\}$ is called {\em a deadlock subset}.
\end{defi}

\begin{thm}\label{T:15}
Suppose $H$ is a conservative set of functions over $D$, $m \in H$ is an arithmetical operation on $B\subseteq \left\{ {\left\{ {x,y} \right\}|x,y \in D ,x \ne y} \right\}$ and a pair
$\phi, \psi \in H$ is a tournament pair on $\overline{B}$.
If an instance of $2-MinHom\left( Inv\left( H \right) \right)$
is arc- and path-consistent, then it cannot be an odd arithmetical deadlock.
\end{thm}

We will begin by introducing some technical concepts from the theory of $CSP$ which we will need in the proof of Theorem \ref{T:15}. An algebra ${\mathbb A}$ is said to be {\em of type $\mathfrak{F}$} if its operations are indexed by elements of the set $\mathfrak{F}$, called terms. For every $f\in \mathfrak{F}$, the corresponding operation is denoted by $f^{{\mathbb A}}$. The universe of an algebra ${\mathbb A_i}$ is denoted by $A_i$.
Recall that $\rho^t = \left\{ {\left( {y,x} \right)|\left( {x,y} \right) \in \rho } \right\}$.

\begin{defi}\label{D:first}
Let a finite set of indexes $I$ be given and every index $i \in I$ corresponds to some algebra ${\mathbb A_i}$ of type $\mathfrak{F}$.
{\em A set of indexed multi-domain predicates over $\left\{ {{\mathbb A_i} } \right\}_{i \in I} $ } is a pair
$\langle {\rho _i } \rangle_{i \in I} ,\langle {\rho _{kl} } \rangle_{k \ne l \in I} $, where for each $i$ and $k \ne l$, $\rho _i $ is a subalgebra of
${\mathbb A_i}$ and $\rho _{kl} $ is a subalgebra of ${\mathbb A_k}  \times \mathbb A_l $. We assume that $\rho _{kl}  = \rho _{lk}^t$.
\end{defi}

\begin{defi}\label{D:first}
A set of indexed multi-domain predicates $\langle {\rho _i } \rangle_{i \in I} ,\langle {\rho _{kl} } \rangle_{k \ne l \in I} $ over $\left\{ {{\mathbb A_i} } \right\}_{i \in I} $
is called {\em arc-consistent} if for distinct $i,j \in I:{\rm{ }}\Pr _1 \rho _{ij}  = \rho _i ,\Pr _2 \rho _{ij}  = \rho _j$.
\end{defi}

\begin{defi}\label{D:first}
A set of indexed multi-domain predicates $\langle {\rho _i } \rangle_{i \in I} ,\langle {\rho _{kl} } \rangle_{k \ne l \in I} $ over $\left\{ {{\mathbb A_i} } \right\}_{i \in I} $
is called {\em path-consistent} if for any distinct $i,j,k \in I:{\rm{ }}\rho _{ik}  \subseteq \rho _{ij}  \circ \rho _{jk} $.
\end{defi}

Introduce the notation $P_i  = \left\{ {\left\{ {x,y} \right\}|x,y \in A_i ,x \ne y} \right\}$.

\begin{defi}\label{D:first}
Assume that algebras $\left\{ {{\mathbb A_i} } \right\}_{i \in I} $ are of type $\mathfrak{F}$, that they are conservative, and
$B_i  \subseteq P_i ,i \in I$.
A term  $m \in \mathfrak{F}$ is called {\em arithmetical on $\left\{ {B_i } \right\}_{i \in I} $}, if for any $i \in I$ $m^{A_i }$ is arithmetical on $B_i$. A pair of terms $\phi, \psi \in \mathfrak{F}$ is called {\em a tournament pair on $\left\{ {B_i } \right\}_{i \in I} $ },
if for any $i \in I$ a pair $\phi ^{A_i }, \psi ^{A_i }$ is a tournament pair on $B_i$.
\end{defi}

We now generalize the notion of an {\em odd arithmetical deadlock} to multi-domain constraints.

\begin{defi}\label{D:first}
Assume that algebras $\left\{ {{\mathbb A_i} } \right\}_{i \in I} $ are of type $\mathfrak{F}$, that they are conservative, and
$B_i  \subseteq P_i ,i \in I$. Furthermore, assume $m \in \mathfrak{F}$ is an arithmetical term on $\left\{ {B_i } \right\}_{i \in I} $ and a pair
$\phi, \psi \in \mathfrak{F}$ is a tournament pair on $\left\{ {P_i /B_i } \right\}_{i \in I} $.
Then, the set of indexed multi-domain predicates $\langle {\rho _i } \rangle_{i \in I} ,\langle {\rho _{kl} } \rangle_{k \ne l \in I} $ over
$\left\{ {{\mathbb A_i} } \right\}_{i \in I} $ is called an {\em odd arithmetical deadlock} if there is a subset
$\left\{ {i_0 ,\dots,i_{n - 1} } \right\} \subseteq I,n \ge 3$ of odd cardinality and
$\left\{ {x_0 ,y_0 } \right\} \in B_{i_0 } ,\dots,\left\{ {x_{n - 1} ,y_{n - 1} } \right\} \in B_{i_{n - 1} } $, such that for
$0\leq k \leq n - 1$:  $\rho _{i{}_k,i{}_{k \oplus 1}}  \cap \left\{ {x_k ,y_k } \right\} \times \left\{ {x_{k \oplus 1} ,y_{k \oplus 1} } \right\} = \begin{picture}(10,10)\end{picture}_{ y_k }^{ x_k }\begin{picture}(10,10)\put(0,10){\line(1,-1){10}}\put(0,0){\line(1,1){10}}\end{picture}^{ x_{k \oplus 1}}_{ y_{k \oplus 1}}$, where $i \oplus j$ denotes $i + j(\bmod {\rm{\,\,}}n)$.
The subset $\left\{ {i_0 ,\dots,i_{n - 1} } \right\}$ is called a {\em deadlock subset}.
\end{defi}

We will now prove the following theorem, which is a generalization of Theorem \ref{T:15}.

\begin{thm}\label{T:152}
Suppose $m \in \mathfrak{F}$ is an arithmetical term on $\left\{ {B_i } \right\}_{i \in I} $, and a pair $\phi, \psi \in \mathfrak{F}$
is a tournament pair on $\left\{ {P_i /B_i } \right\}_{i \in I} $. If a set of indexed multi-domain predicates
$\langle {\rho _i } \rangle_{i \in I} ,\langle {\rho _{kl} } \rangle_{k \ne l \in I} $ over $\left\{ {{\mathbb A_i} } \right\}_{i \in I} $
is arc- and path-consistent, then it cannot be an odd arithmetical deadlock.
\end{thm}

Any instance of $2-MinHom\left( Inv\left( H \right) \right)$ can be considered as a set of indexed multi-domain predicates
over $\left\{ {{\mathbb A_i} } \right\}_{i \in I} $ where $I$ is a set of variables and $\mathbb A_i = \mathbb A$.
By defining $B_i = B$ we see that Theorem \ref{T:15} is a special case of Theorem \ref{T:152}.
Before proving Theorem \ref{T:152}, we need to prove some preliminary lemmas.

Recall that a congruence of an algebra ${\mathbb A}$ is an equivalence relation on $A$ that is a subalgebra of ${\mathbb A}^2$.
If $\theta $ is a congruence of ${\mathbb A}$ and $a \in A$, then equivalence class of $\theta $ containing $a$ is denoted by
$a^\theta$.
If for each $s \in I$, $\theta _s $ is a congruence of ${\mathbb A_s} $, then
$\rho _i /\theta _i  = \left\{ {x^{\theta _i } |x \in \rho _i } \right\}$ and $\rho _{kl} /\left( {\theta _k  \times \theta _l } \right) = \left\{ {\left( {x^{\theta _k } ,y^{\theta _l } } \right)|\left( {x,y} \right) \in \rho _{kl} } \right\}$,
which we view as subalgebras of ${\mathbb A_i} /\theta _i $ and $\left( {{\mathbb A_k} /\theta _k } \right) \times \left( {\mathbb A_l /\theta _l } \right)$.

\begin{lem}\label{L:3}
Let $\theta _i $ be a congruence of ${\mathbb A_i}$ for each $i \in I$ and assume that
a set of indexed multi-domain predicates
$\langle {\rho _i } \rangle_{i \in I} ,\langle {\rho _{kl} } \rangle_{k \ne l \in I} $ over $\left\{ {{\mathbb A_i} } \right\}_{i \in I} $
is arc- and path-consistent. Then a set of indexed multi-domain predicates
$\left\{ {\rho _i /\theta _i } \right\}_{i \in I} ,\left\{ {\rho _{kl} /\left( {\theta _k  \times \theta _l } \right)} \right\}_{k \ne l \in I} $
over $\left\{ {A_i /\theta _i } \right\}_{i \in I} $ is arc- and path-consistent, too.
\end{lem}

\proof[Proof.]
Let $n_i :{\mathbb A_i}  \to {\mathbb A_i} /\theta _i $ be natural homomorphisms, i.e., $n_i(x) = x^{\theta_i}$.
Obviously, $\rho _i /\theta _i  = \left\{ {n_i \left( x \right)|x \in \rho _i } \right\},\rho _{kl} /\left( {\theta _k  \times \theta _l } \right) = \left\{ {\left( {n_k \left( x \right),n_l \left( y \right)} \right)|\left( {x,y} \right) \in \rho _{kl} } \right\}$ and $\Pr _1 \left[ {\rho _{kl} /\left( {\theta _k  \times \theta _l } \right)} \right] = \left\{ {n_k \left( x \right)|x \in \Pr _1 \rho _{kl} } \right\} = \Pr _1 \rho _{kl} /\theta _k $. Analogously, we can prove that $\Pr _2 \left[ {\rho _{kl} /\left( {\theta _k  \times \theta _l } \right)} \right] = \Pr _2 \rho _{kl} /\theta _l $.

From arc-consistency it follows that $\Pr _1 \rho _{kl}  = \rho _k ,\Pr _2 \rho _{kl}  = \rho _l $, and we have
$\Pr _1 \left[ {\rho _{kl} /\left( {\theta _k  \times \theta _l } \right)} \right] = \rho _k /\theta _k ,\Pr _2 \left[ {\rho _{kl} /\left( {\theta _k  \times \theta _l } \right)} \right] = \rho _l /\theta _l $.
This is equivalent to arc-consistency of the set $\left\{ {\rho _i /\theta _i } \right\}_{i \in I} ,\left\{ {\rho _{kl} /\left( {\theta _k  \times \theta _l } \right)} \right\}_{k \ne l \in I} $.

The path-consistency condition $\rho _{ik}  \subseteq \rho _{ij}  \circ \rho _{jk} $ gives us:
\[
\begin{array}{l}
\rho _{ij} /\left( {\theta _i  \times \theta _j } \right) \circ \rho _{jk} /\left( {\theta _j  \times \theta _k } \right) = \\
= \left\{ {\left( {n_i \left( x \right),n_j \left( y \right)} \right)|\left( {x,y} \right) \in \rho _{ij} } \right\} \circ \left\{ {\left( {n_j \left( z \right),n_k \left( t \right)} \right)|\left( {z,t} \right) \in \rho _{jk} } \right\} \supseteq \\
\supseteq \left\{ {\left( {n_i \left( x \right),n_k \left( t \right)} \right)|\left( {x,y} \right) \in \rho _{ij} ,\left( {y,t} \right) \in \rho _{jk} } \right\} \supseteq \\
\supseteq \left\{ {\left( {n_i \left( x \right),n_k \left( t \right)} \right)|\left( {x,t} \right) \in \rho _{ik} } \right\} = \rho _{ik} /\left( {\theta _i  \times \theta _k } \right) \\
\end{array}
\]

This is equivalent to path-consistency of
$\left\{ {\rho _i /\theta _i } \right\}_{i \in I}$ and $\left\{ {\rho _{kl} /\left( {\theta _k  \times \theta _l } \right)} \right\}_{k \ne l \in I} $.
\qed

For $\rho  \subseteq A_1  \times A_2 $, let
$\rho \left( {x, \cdot } \right) = \left\{ {y|\rho \left( {x,y} \right)} \right\}$ and $\rho \left( { \cdot ,x} \right) = \left\{ {y|\rho \left( {y,x} \right)} \right\}$.

\begin{lem}\label{L:4}
Suppose algebras $\left\{ {{\mathbb A_i} } \right\}_{i = 1,2} $ of type $\mathfrak{F}$ are conservative and  $B_i  \subseteq P_i ,i = 1,2$.
Furthermore, assume that $m \in \mathfrak{F}$
is an arithmetical term on $B_i ,i = 1,2$, and a pair $\phi, \psi \in \mathfrak{F}$ is a tournament pair on $P_i /B_i ,i = 1,2$.
If $\rho $ is a subalgebra of ${\mathbb A_1}  \times {\mathbb A_2} $ and there are $\left\{ {x_i ,y_i } \right\} \in B_i ,i = 1,2$, such that
$\rho  \cap \left\{ {x_1 ,y_1 } \right\} \times \left\{ {x_2 ,y_2 } \right\} =\begin{picture}(10,10)\end{picture}_{ y_1 }^{ x_1}\begin{picture}(10,10)\put(0,10){\line(1,-1){10}}\put(0,0){\line(1,1){10}}\end{picture}^{ x_2}_{ y_2}$,
then $\rho \left( {x_1 , \cdot } \right) \cap \rho \left( {y_1 , \cdot } \right) = \emptyset $ and $\rho \left( { \cdot ,x_2 } \right) \cap \rho \left( { \cdot ,y_2 } \right) = \emptyset $.
\end{lem}

\proof[Proof.] Suppose, for example, that $t \in \rho \left( {x_1 , \cdot } \right) \cap \rho \left( {y_1 , \cdot } \right)$. Then, if $\left\{ {x_2 ,t} \right\} \in B_2 $, we have:

$\left( {\begin{array}{*{20}c}
   {x_1 }  \\
   t  \\
\end{array}} \right),\left( {\begin{array}{*{20}c}
   {y_1 }  \\
   t  \\
\end{array}} \right),\left( {\begin{array}{*{20}c}
   {y_1 }  \\
   {x_2 }  \\
\end{array}} \right) \in \rho  \Rightarrow \left( {\begin{array}{*{20}c}
   {m^{{\mathbb A_1} } \left( {x_1 ,y_1 ,y_1 } \right)}  \\
   {m^{{\mathbb A_2} } \left( {t,t,x_2 } \right)}  \\
\end{array}} \right) = \left( {\begin{array}{*{20}c}
   {x_1 }  \\
   {x_2 }  \\
\end{array}} \right) \in \rho {\rm{  }}$

If $\left\{ {x_2 ,t} \right\} \in P_2 /B_2 $, then there is a $\lambda \in \mathfrak{F}: \mathop  \downarrow \limits_{x_2 }^t \lambda ^{{\mathbb A_2} } $
where either $\lambda  = \phi $ or $\lambda  = \psi $ and we have:

$\left( {\begin{array}{*{20}c}
   {x_1 }  \\
   t  \\
\end{array}} \right),\left( {\begin{array}{*{20}c}
   {y_1 }  \\
   {x_2 }  \\
\end{array}} \right) \in \rho  \Rightarrow \left( {\begin{array}{*{20}c}
   {\lambda ^{{\mathbb A_1} } \left( {x_1 ,y_1 } \right)}  \\
   {\lambda ^{{\mathbb A_2} } \left( {t,x_2 } \right)}  \\
\end{array}} \right) = \left( {\begin{array}{*{20}c}
   {x_1 }  \\
   {x_2 }  \\
\end{array}} \right) \in \rho {\rm{  }}$

Now we see that $\rho \left( {x_1 , \cdot } \right) \cap \rho \left( {y_1 , \cdot } \right) = \emptyset $
(analogously $\rho \left( { \cdot ,x_2 } \right) \cap \rho \left( { \cdot ,y_2 } \right) = \emptyset $).
\qed

For $\rho  \subseteq A_1  \times A_2 $, $\theta _1^\rho$ and $\theta _2^\rho$ denote the transitive closures of $\rho \circ \rho^t$ and
$\rho^t \circ \rho$ respectively.

\begin{lem}\label{L:5}
Suppose algebras $\left\{ {{\mathbb A_i} } \right\}_{i = 1,2} $ of type $\mathfrak{F}$ are conservative and $B_i  \subseteq P_i ,i = 1,2$.
Suppose also that $m \in \mathfrak{F}$ is arithmetical term on $B_i ,i = 1,2$, and a pair $\phi, \psi \in \mathfrak{F}$ is a tournament pair on $P_i /B_i ,i = 1,2$.
If $\rho $ is a subalgebra of ${\mathbb A_1}  \times {\mathbb A_2} $ and there are $\left\{ {x_i ,y_i } \right\} \in B_i ,i = 1,2$, such that
$\rho  \cap \left\{ {x_1 ,y_1 } \right\} \times \left\{ {x_2 ,y_2 } \right\} =\begin{picture}(10,10)\end{picture}_{ y_1 }^{ x_1 }\begin{picture}(10,10)\put(0,10){\line(1,-1){10}}\put(0,0){\line(1,1){10}}\end{picture}^{ x_2}_{ y_2}$,
then $x_i^{\theta _i^\rho  }  \ne y_i^{\theta _i^\rho  } ,i = 1,2$.
\end{lem}

\proof[Proof.]
Note that for $x \in A_1 $, the equivalence class $x^{\theta _1^\rho  } $ can be obtained by the following procedure:
$U_1  = \left\{ x \right\}$, $U_2  = \left\{ {t|\exists y \in U_1 {\rm{\,\,}}\rho \left( {y,t} \right)} \right\}$, $U_3  = \left\{ {t|\exists y \in U_2 {\rm{\,\,}}\rho \left( {t,y} \right)} \right\}$, $U_4  = \left\{ {t|\exists y \in U_3 {\rm{\,\,}}\rho \left( {y,t} \right)} \right\}$ and so on.
The resulting equivalence class is $U_1  \cup U_3  \cup U_5 \dots$.
Consider this process for elements $x_1 ,y_1 $ and denote the corresponding sets by $U_1^{x_1 } ,U_2^{x_1 } ,\dots$ and $U_1^{y_1 } ,U_2^{y_1 } ,\dots$.
We prove by induction that
$U_s^{x_1 }  \cap U_s^{y_1 }  = \emptyset $ and
$\delta _k  \buildrel \Delta \over = \left( {U_k^{x_1 } } \right)^2  \cup \left( {U_k^{y_1 } } \right)^2 $ is a congruence of
${\mathbb A_1} |_{U_k^{x_1 }  \cup U_k^{y_1 } } $, if $k$ is odd, or of ${\mathbb A_2} |_{U_k^{x_1 }  \cup U_k^{y_1 } } $, if  $k$ is even.

Base of induction. Obviously, $U_1^{x_1 }  \cap U_1^{y_1 }  = \emptyset $.
Since $\rho' = \rho  \cap \left\{ {x_1 ,y_1 } \right\} \times \left\{ {x_2 ,y_2 } \right\} =\begin{picture}(10,10)\end{picture}_{ y_1 }^{ x_1 }\begin{picture}(10,10)\put(0,10){\line(1,-1){10}}\put(0,0){\line(1,1){10}}\end{picture}^{ x_2}_{ y_2}$
is a subalgebra of ${\mathbb A_1} |_{\left\{ {x_1 ,y_1 } \right\}}  \times {\mathbb A_2} |_{\left\{ {x_2 ,y_2 } \right\}} $, we see that $\left( {U_1^{x_1 } } \right)^2  \cup \left( {U_1^{y_1 } } \right)^2  = \theta _1^{\rho '} $
is a congruence of ${\mathbb A_1} |_{\left\{ {x_1 ,y_1 } \right\}} $.

Suppose the assertion is true for $s \le k$.
Consider the case when $k$ is even (the odd case is analogous).
Let $\rho ' = \rho  \cap \left( {A_1  \times \left( {U_k^{x_1 }  \cup U_k^{y_1 } } \right)} \right)$.
Clearly, $\rho '/\left( { = ^{A_1 }  \times \delta _k } \right)$ is a subalgebra of
${\mathbb A_1}  \times \left( {{\mathbb A_2} |_{U_k^{x_1 }  \cup U_k^{y_1 } } /\delta _k } \right)$ and from $y_2  \in U_k^{x_1 } ,x_2  \in U_k^{y_1 } $
we have
\[
\begin{array}{l}
 U_{k + 1}^{x_1 }  = \rho '/\left( { = ^{A_1 }  \times \delta _k } \right)\left( { \cdot ,y_2^{\delta _k } } \right) \\
 U_{k + 1}^{y_1 }  = \rho '/\left( { = ^{A_1 }  \times \delta _k } \right)\left( { \cdot ,x_2^{\delta _k } } \right) \\
 \end{array}
\]

A pair of algebras ${\mathbb A_1} ,{\mathbb A_2} |_{U_k^{x_1 }  \cup U_k^{y_1 } } /\delta _k $ of type $\mathfrak{F}$ obviously
satisfy conditions of Lemma \ref{L:4}.
Since $\rho \left( {x_1 , \cdot } \right) \subseteq U_k^{x_1 }$ and $\rho \left( {y_1 , \cdot } \right) \subseteq U_k^{y_1 } $,
we have
\[
\rho '/\left( { = ^{A_1 }  \times \delta _k } \right) \cap \left\{ {x_1 ,y_1 } \right\} \times \left\{ {x_2^{\delta _k } ,y_2^{\delta _k } } \right\} =\begin{picture}(10,10)\end{picture}_{ y_1 }^{ x_1 }\begin{picture}(10,10)\put(0,10){\line(1,-1){10}}\put(0,0){\line(1,1){10}}\end{picture}^{ x_2^{\delta _k }}_{ y_2^{\delta _k }}.
\]
From Lemma \ref{L:4} we see that
\[
\rho '/\left( { = ^{A_1 }  \times \delta _k } \right)\left( { \cdot ,y_2^{\delta _k } } \right) \cap \rho '/\left( { = ^{A_1 }  \times \delta _k } \right)\left( { \cdot ,x_2^{\delta _k } } \right) = \emptyset
\]
which is equivalent to $U_{k + 1}^{x_1 }  \cap U_{k + 1}^{y_1 }  = \emptyset $.

From the emptiness of this intersection, we conclude that the predicate
$\sigma  = \theta^{\rho '/\left( { = ^{A_1 }  \times \delta _k } \right)}_1$ is a congruence and equals to
$\left( {U_{k + 1}^{x_1 } } \right)^2  \cup \left( {U_{k + 1}^{y_1 } } \right)^2 $, and the induction is completed.
\qed

\begin{lem}\label{L:6}
Suppose ${\mathbb A}$ is three-element algebra containing an operation $h:A^3  \to A$ that is arithmetical on $\left\{\{a,b\}|a,b\in A, a\ne b\right\}$.
Then, there cannot be two different nontrivial(i.e. not equal to $A^2 $ or $ = ^A $) congruences of this algebra.
\end{lem}

\proof[Proof.]
We give a proof by contradiction.
Without loss of generality we can assume that $A = \left\{ {0,1,2} \right\}$ and
$ \sim ^1  = \left\{ {\left( {0,0} \right),\left( {1,1} \right),\left( {2,2} \right),\left( {0,1} \right)} \right\}$,
$ \sim ^2  = \left\{ {\left( {0,0} \right),\left( {1,1} \right),\left( {2,2} \right),\left( {1,2} \right)} \right\}$.
Since $h$ preserve $ \sim ^1 $, we have:
\[
\begin{array}{*{20}c}
  {h\left( {1,1,2} \right) = 2}  \\
  {h\left( {0,1,2} \right) = ?}  \\
\end{array} \Rightarrow h\left( {0,1,2} \right) = 2.
\]
Preservation of $ \sim ^2 $ leads to contradiction:
\[
\begin{array}{*{20}c}
   {h\left( {0,1,1} \right) = 0}  \\
   {h\left( {0,1,2} \right) = ?}  \\
\end{array} \Rightarrow h\left( {0,1,2} \right) = 0.
\]
\qed

\proof[Proof of Theorem \ref{T:152}] Suppose to the contrary that there exists a set of indexed multi-domain predicates that is an odd arithmetical deadlock. We can assume that $I = \left\{ {0,\dots,2d} \right\}$ and
$\left\{ {x_0 ,y_0 } \right\} \in B_0 ,\dots,\left\{ {x_{2d} ,y_{2d} } \right\} \in B_{2d} $, such that
$\rho _{k,k \oplus 1}  \cap \left\{ {x_k ,y_k } \right\} \times \left\{ {x_{k \oplus 1} ,y_{k \oplus 1} } \right\} = \begin{picture}(10,10)\end{picture}_{ y_k}^{ x_k}\begin{picture}(10,10)\put(0,10){\line(1,-1){10}}\put(0,0){\line(1,1){10}}\end{picture}^{ x_{k \oplus 1}}_{ y_{k \oplus 1}} $, where $i \oplus j$ denotes $i + j(\bmod {\rm{\,\,}}2d+1)$.

Consider the predicates $\rho _{k \ominus 1,k}$ and $\rho _{k,k \oplus 1} $. Let $\theta-$ and $\theta+$ denote congruences $\theta _2^{\rho _{k \ominus 1, k} }, \theta _1^{\rho _{k,k \oplus 1} } $ consistently.
By Lemma \ref{L:5}, $x_k^{\theta+ }  \ne y_k^{\theta+ } $.
Obviously, $\rho _{k,k \oplus 1} \left( { \cdot ,x_{k \oplus 1} } \right) \subseteq y_k^{\theta+ } $ and
$\rho _{k,k \oplus 1} \left( { \cdot ,y_{k \oplus 1} } \right) \subseteq x_k^{\theta+ } $.
Therefore, we conclude that
\[
\rho _{k,k \oplus 1} /\left( {\left(\theta+\right)  \times \left( = ^{A_{k \oplus 1} }\right) } \right)
\cap \left\{ {x_k^{\theta+ } ,y_k^{\theta+ } } \right\} \times \left\{ {x_{k \oplus 1} ,y_{k \oplus 1} } \right\}
= \begin{picture}(10,10)\end{picture}_{ y_k^{\theta+ }}^{ x_k^{\theta+ }}\begin{picture}(10,10)\put(0,10){\line(1,-1){10}}\put(0,0){\line(1,1){10}}\end{picture}^{ x_{k \oplus 1}}_{ y_{k \oplus 1}}.
\]

Let us show that $\rho _{k \ominus 1,k} \left( {x_{k \ominus 1} , \cdot } \right) \cap x_k^{\theta+ }  = \emptyset $ and
$\rho _{k \ominus 1,k} \left( {y_{k \ominus 1} , \cdot } \right) \cap y_k^{\theta+ }  = \emptyset $.
Suppose to the contrary that the first one is false (the other case is absolutely analogous), i.e.
$t \in \rho _{k \ominus 1,k} \left( {x_{k \ominus 1} , \cdot } \right) \cap x_k^{\theta+ } $.
From $\rho _{k \ominus 1,k} \left( {x_{k \ominus 1} ,t} \right)$, we see that
$\left( {t,y_k } \right) \in \theta- $.
But, from $t \in x_k^{\theta+ } $, we conclude that
$\left( {t,x_k } \right) \in \theta+ $.
Consider the three-element algebra
${\mathbb A_k} |_{\left\{ {x_k ,y_k ,t} \right\}} $. The congruences
$\theta+ $, $\theta- $
restricted to that algebra are equal to $\left\{ {\left\{ {x_k ,t} \right\},\left\{y_k\right\}} \right\}$ and $\left\{ {\left\{ {y_k ,t} \right\},\left\{x_k\right\}} \right\}$, since, by Lemma \ref{L:5},
$x_k^{\theta+ }  \ne y_k^{\theta+ } $ and
$x_k^{\theta- }  \ne y_k^{\theta- } $.
It is easy to see that the three-element conservative algebra ${\mathbb A_k} |_{\left\{ {x_k ,y_k ,t} \right\}} $ with $\left\{ {x_k ,y_k } \right\} \in B_k$
has such congruences only if
$m$ is arithmetical on $\left\{\{x_k ,y_k\}, \{y_k ,t\},\{x_k,t\}\right\}$.
This contradicts Lemma \ref{L:6}.

From $\rho _{k \ominus 1,k} \left( {x_{k \ominus 1} , \cdot } \right) \cap x_k^{\theta+ }  = \emptyset $ and
$\rho _{k \ominus 1,k} \left( {y_{k \ominus 1} , \cdot } \right) \cap y_k^{\theta+ }  = \emptyset $,
we conclude that
\[
\rho _{k \ominus 1,k} /\left( {\left( = ^{A_{k \ominus 1} }\right)  \times \left(\theta+\right)  } \right)
\cap\left\{ {x_{k \ominus 1} ,y_{k \ominus 1} } \right\} \times \left\{ {x_k^{\theta+ } ,y_k^{\theta+ } } \right\}
=\begin{picture}(10,10)\end{picture}_{ y_{k \ominus 1} }^{ x_{k \ominus 1} }\begin{picture}(10,10)\put(0,10){\line(1,-1){10}}\put(0,0){\line(1,1){10}}\end{picture}^{ x_k^{\theta+ }}_{ y_k^{\theta+ }}.
\]
Therefore, changing a system of one-type algebras $\left\{ {{\mathbb A_i} } \right\}_{i \in I}$ to $\left\{ {{\mathbb A_i} /\lambda _i } \right\}_{i \in I} $ where
\[
\lambda _i  = \left\{ {\begin{array}{*{20}c}
   {\theta _1^{\rho _{k,k \oplus 1} } ,{\rm{if\,\,}}i = k}  \\
   { = ^{A_i } ,{\rm{otherwise\,\,\,}}}  \\
\end{array}} \right.
\]
we obtain, by Lemma \ref{L:3}, an arc- and path-consistent set of indexed predicates
$\left\{ {\rho _i /\lambda _i } \right\}_{i \in I} ,\left\{ {\rho _{kl} /\left( {\lambda _k  \times \lambda _l } \right)} \right\}_{k \ne l \in I} $.
The resulting set of predicates will be an odd arithmetical deadlock, too.

Analogously, we can prove that changing a system of one-type algebras $\left\{ {{\mathbb A_i} } \right\}_{i \in I}$ to
$\left\{ {{\mathbb A_i} /\lambda _i } \right\}_{i \in I} $, where
\[
\lambda _i  = \left\{ {\begin{array}{*{20}c}
   {\theta _2^{\rho _{k \ominus 1, k} } ,{\rm{if\,\,}}i = k}  \\
   { = ^{A_i } ,{\rm{otherwise\,\,\,}}}  \\
\end{array}} \right.
\]
result in an arc- and path-consistent set of indexed predicates
$\left\{ {\rho _i /\lambda _i } \right\}_{i \in I} ,\left\{ {\rho _{kl} /\left( {\lambda _k  \times \lambda _l } \right)} \right\}_{k \ne l \in I} $,
which will be an odd arithmetical deadlock.

By using those transformations for different $k$ successively, we eventually obtain an arc- and path-consistent
$\left\{ {\rho '_i } \right\}_{i \in I} ,\left\{ {\rho '_{kl} } \right\}_{k \ne l \in I} $, such that
$\forall k{\rm{\,\,}}\rho '_{k,k \oplus 1}  \cap \left\{ {x'_k ,y'_k } \right\} \times \left\{ {x'_{k \oplus 1} ,y'_{k \oplus 1} } \right\} = \begin{picture}(10,10)\end{picture}_{ y'_k }^{ x'_k }\begin{picture}(10,10)\put(0,10){\line(1,-1){10}}\put(0,0){\line(1,1){10}}\end{picture}^{ x'_{k \oplus 1}}_{ y'_{k \oplus 1} }$ and $\forall k{\rm{\,\,}}\rho _{k,k \oplus 1} \left( { \cdot ,x'_{k \oplus 1} } \right) = \left\{ {y'_k } \right\},\rho _{k,k \oplus 1} \left( { \cdot ,y'_{k \oplus 1} } \right) = \left\{ {x'_k } \right\}, \rho _{k \ominus 1,k} \left( {x'_{k \ominus 1} , \cdot } \right) = \left\{ {y'_k } \right\}$ and $\rho _{k \ominus 1,k} \left( {y'_{k \ominus 1} , \cdot } \right) = \left\{ {x'_k } \right\}$.
We show that there is no such set.

From path-consistency we conclude that for any $0 \le k < l \le 2d$:
$\rho '_{kl}  \subseteq \rho '_{k,k + 1}  \circ \rho '_{k + 1,k + 2}  \circ \dots \circ \rho '_{l - 1,l} $. Hence,
\[
\rho '_{k,k + 1}  \circ \rho '_{k + 1,k + 2}  \circ \dots \circ \rho '_{l - 1,l} \left( {x'_k , \cdot } \right) = \left\{ {\begin{array}{*{20}c}
   {\left\{ {x'_l } \right\},{\rm{if\,\,}}l - k \rm{\,\,even}}  \\
   {\left\{ {y'_l } \right\},{\rm{if\,\,}}l - k \rm{\,\,odd }}  \\
\end{array}} \right.
\]

Since $\rho '_{kl} \left( {x'_k , \cdot } \right)$ is not empty, we see that
\[
\rho '_{kl} \left( {x'_k , \cdot } \right) = \left\{ {\begin{array}{*{20}c}
   {\left\{ {x'_l } \right\},{\rm{if\,\,}}l - k \rm{\,\,even}}  \\
   {\left\{ {y'_l } \right\},{\rm{if\,\,}}l - k \rm{\,\,odd }}  \\
\end{array}} \right.
\]

However, we have $\rho '_{0,2d}  \cap \left\{ {x'_0 ,y'_0 } \right\} \times \left\{ {x'_{2d} ,y'_{2d} } \right\} = \begin{picture}(10,10)\end{picture}_{ y'_0 }^{ x'_0 }\begin{picture}(10,10)\put(0,10){\line(1,-1){10}}\put(0,0){\line(1,1){10}}\end{picture}^{ x'_{2d}}_{ y'_{2d } }$
which contradicts that $\rho '_{0,2d} \left( {x'_0 , \cdot } \right) = \left\{ {x'_{2d} } \right\}$.
\qed

\section{Final step in a proof of polynomial case}

\begin{thm}\label{T:16}
Suppose that $F$ satisfies the necessary local conditions and that the graph $T_F = \left( {M^o ,P} \right)$ is bipartite. Then for every path- and arc-consistent instance of $2 - MinHom\left( {Inv\left( {F} \right)} \right)$, its microstructure graph forbids subgraphs of type $S_{2p + 1} ,p \ge 2$.
\end{thm}

\proof[Proof.]
Suppose to the contrary that we have a arc- and path-consistent instance $I = \left( X, U, B, w\right)$ of $2-MinHom\left( Inv\left( F \right) \right)$ with constraints pair $U = \langle {\rho _i } \rangle_{1 \le i \le n}$, $B = \langle {\rho _{kl} } \rangle_{1 \le k \ne l \le n} $
and its microstructure graph has a subgraph of type $S_{2p + 1} ,p \ge 2$.
For convenience, let us introduce  $\rho _{ii}  = \left\{ {\left( {a,a} \right)|a \in \rho _i } \right\}$.
Then, there is a set of pairs $\left\{ {\left( {i_0 ,b_0 } \right),\left( {i_1 ,b_1 } \right),\dots,\left( {i_{2p} ,b_{2p} } \right)} \right\}$,
such that for $0\leq l \leq 2p $: $\left( {b_l ,b_{l \oplus 1} } \right) \notin \rho _{i_l i_{l \oplus 1} } $ and
$\left( {b_l ,b_{l \oplus 2} } \right) \in \rho _{i_l i_{l \oplus 2} } $, where $i \oplus j$ denotes $i + j(\bmod {\rm{\,\,}}2p+1)$.

From $\left( {b_l ,b_{l \oplus 2} } \right) \in \rho _{i_l i_{l \oplus 2} } $ and the path-consistency condition $\rho _{i_l i_{l \oplus 2} }  \subseteq \rho _{i_l i_{l \oplus 1} }  \circ \rho _{i_{l \oplus 1} i_{l \oplus 2} } $,
we see that there is $a_{l \oplus 1} $, such that
$\left( {b_l ,a_{l \oplus 1} } \right) \in \rho _{i_l i_{l \oplus 1} } $ and
$\left( {a_{l \oplus 1} ,b_{l \oplus 2} } \right) \in \rho _{i_{l \oplus 1} i_{l \oplus 2} } $.

Consider the predicate $\rho '_{l,l \oplus 1}  = \rho _{i_l i_{l \oplus 1} }  \cap \left\{ {a_l ,b_l } \right\} \times \left\{ {a_{l \oplus 1} ,b_{l \oplus 1} } \right\} \in Inv\left( F \right)$.
Obviously, $\rho '_{l,l \oplus 1} $ equals to either
$ \begin{picture}(10,10)\end{picture}_{ b_l}^{ a_l}\begin{picture}(10,10)\put(0,10){\line(1,-1){10}}\put(0,0){\line(1,1){10}}\put(0,10){\line(1,0){10}}\end{picture}^{a_{l \oplus 1}}_{ b_{l \oplus 1}}$ or
$ \begin{picture}(10,10)\end{picture}_{ b_l}^{ a_l}\begin{picture}(10,10)\put(0,10){\line(1,-1){10}}\put(0,0){\line(1,1){10}}\end{picture}^{a_{l \oplus 1}}_{ b_{l \oplus 1}}$.

Let us show that if $\left\{ {a_l ,b_l } \right\} \in \overline M $, then
$\left\{ {a_{l \oplus 1} ,b_{l \oplus 1} } \right\} \in \overline M $, too.
Assume to the contrary that $\left\{ {a_{l \oplus 1} ,b_{l \oplus 1} } \right\} \in M$.
Then, by Theorem \ref{T:11}, there is a $\phi  \in F:\mathop  \downarrow \limits_{b_{l \oplus 1} }^{a_{l \oplus 1} } \phi$, where
$\phi |_{\left\{ {a_l ,b_l } \right\}} $ is a projection on the first coordinate.
In this case, $\phi $ preserves neither $ \begin{picture}(10,10)\end{picture}_{ b_l}^{ a_l}\begin{picture}(10,10)\put(0,10){\line(1,-1){10}}\put(0,0){\line(1,1){10}}\put(0,10){\line(1,0){10}}\end{picture}^{a_{l \oplus 1}}_{ b_{l \oplus 1}}$
nor $ \begin{picture}(10,10)\end{picture}_{ b_l}^{ a_l}\begin{picture}(10,10)\put(0,10){\line(1,-1){10}}\put(0,0){\line(1,1){10}}\end{picture}^{a_{l \oplus 1}}_{ b_{l \oplus 1}}$, because
\[
\left( {\begin{array}{*{20}c}
   {b_l }  \\
   {b_{l \oplus 1} }  \\
\end{array}} \right) = \left( {\begin{array}{*{20}c}
   {\phi \left( {b_l ,a_l } \right)}  \\
   {\phi \left( {a_{l \oplus 1} ,b_{l \oplus 1} } \right)}  \\
\end{array}} \right).
\]

Hence, we need to consider two cases only: 1) $\forall l{\rm{\,\,}}\left\{ {a_l ,b_l } \right\} \in M$ and 2) $\forall l{\rm{\,\,}}\left\{ {a_l ,b_l } \right\} \in \overline M $.
In the first case, we have
$\left\langle {\left( {a_l ,b_l } \right),\left( {a_{l \oplus 1} ,b_{l \oplus 1} } \right)} \right\rangle  \in P$,
i.e., there is an odd cycle in $T_F$ which contradicts that $T_F$ is bipartite.

Now, consider the case $\forall l{\rm{\,\,}}\left\{ {a_l ,b_l } \right\} \in \overline M $.
By Theorem \ref{T:12}, there is a function $m\in F$, arithmetical on $\overline M$.
If  $\rho '_{l,l \oplus 1}  = \begin{picture}(10,10)\end{picture}_{ b_l}^{ a_l}\begin{picture}(10,10)\put(0,10){\line(1,-1){10}}\put(0,0){\line(1,1){10}}\put(0,10){\line(1,0){10}}\end{picture}^{a_{l \oplus 1}}_{ b_{l \oplus 1}}$, then we have that
\[
\left( {\begin{array}{*{20}c}
   {b_l }  \\
   {b_{l \oplus 1} }  \\
\end{array}} \right) = \left( {\begin{array}{*{20}c}
   {m\left( {a_l ,a_l ,b_l } \right)}  \\
   {m\left( {b_{l \oplus 1} ,a_{l \oplus 1} ,a_{l \oplus 1} } \right)} \\
\end{array}} \right)
\in \rho '_{l,l \oplus 1}
\]
and $\rho '_{l,l \oplus 1} = \begin{picture}(10,10)\end{picture}_{ b_l}^{ a_l}\begin{picture}(10,10)\put(0,10){\line(1,-1){10}}\put(0,0){\line(1,1){10}}\end{picture}^{a_{l \oplus 1}}_{ b_{l \oplus 1}}$.

Consider the set $\left\{ {i_0,i_1,\dots,i_{2p}} \right\}$. Suppose first that all $i_0,i_1,\dots,i_{2p}$ are distinct.
Then, Theorems \ref{T:11} and \ref{T:12} show us that we have an arithmetical operation $m\in F$ on $\overline M$ and a tournament pair $\phi, \psi \in F$ on $M$. It is easy to see that an instance of $2-MinHom\left( Inv\left( F \right) \right)$ with constraints pair $U = \langle {\rho _i } \rangle_{1 \le i \le n} , B = \langle {\rho _{kl} } \rangle_{1 \le k \ne l \le n} $ is an odd arithmetical deadlock where $\left\{ {i_0,i_1,\dots,i_{2p}} \right\}$
is a deadlock set. This contradicts that $I$ is arc- and path-consistent.

The case when the elements $i_0,i_1,\dots,i_{2p}$ are not distinct can be reduced to the previous case by the following trick:
introduce a new set of variables $X' = \left\{ {\left( {i_0 ,0} \right),\left( {i_1 ,1} \right),\dots,\left( {i_{2p} ,2p} \right)} \right\}$
and $\rho_{\left( {i_s ,s} \right)} = \rho_{i_s}$, where $0\leq s\leq 2p$. If $i_m \ne i_n$, then
$\rho _{\left( {i_m ,m} \right), \left( {i_n ,n} \right)} = \rho _{i_m, i_n}$, else $\rho _{\left( {i_m ,m} \right), \left( {i_n ,n} \right)} =
\left\{ {(a,a) | a \in \rho _{i_m}} \right\}$.
It is easy to see that an instance with constraints pair $U = \left\{ {\rho _i } \right\}_{i\in X'} , B = \left\{ {\rho _{kl} } \right\}_{k \ne l\in X'} $
satisfy the conditions of Theorem \ref{T:15} and is an odd arithmetical deadlock, where the set $\left\{ {\left( {i_0 ,0} \right),\left( {i_1 ,1} \right),\dots,\left( {i_{2p} ,2p} \right)} \right\}$
is a deadlock set.
Therefore, we have a contradiction.
\qed

\proof[Proof of polynomial case of Theorem \ref{T:7}.] The conditions of Theorem \ref{T:7} coincides with the conditions of Theorem \ref{T:16} so
the microstructure graph of an arc- and path-consistent instance forbids subgraphs of type
$S_{2p + 1} ,p \ge 2$. By Theorem \ref{T:14}, it is perfect and, by Theorem \ref{T:perfect}, we see that the class $F$ is tractable.
\qed

Theorems \ref{T:6} and \ref{T:7} give the required dichotomy for conservative algebras, which implies the dichotomy for conservative constraint languages.
By Theorem \ref{T:3}, we have the following general dichotomy.

\begin{thm}\label{T:17}
If $MinHom\left( \Gamma \right)$ is not tractable then it is NP-hard.
\end{thm}

\section{Tractable constraint languages}
It is possible to reformulate our results in terms of constraint languages. Let ${\sf lin}_{a_0,a_1}$ denote the predicate
$\left\{(a_x,a_y,a_z)| x,y,z\in \left\{0,1\right\}, x\oplus y\oplus z = 0\right\}$ where $\oplus$ denotes an addition modulo 2. For example,
${\sf lin}_{0,1} = \left\{(x,y,z)| x,y,z\in \left\{0,1\right\}, x\oplus y\oplus z = 0\right\}$.

\begin{thm}\label{T:18}
Suppose $\Gamma$ is a constraint language over $A$ which is a conservative relational clone, then either
\begin{itemize}
   \item $\exists\,\,a\ne b\in A$ such that $\begin{picture}(10,10)\end{picture}_{b}^{a}\begin{picture}(10,10)\put(0,10){\line(1,-1){10}}\put(0,0){\line(1,1){10}}\put(0,10){\line(1,0){10}}\end{picture}_{b}^{a} \in \Gamma$, or
   \item $\exists\,\,a\ne b\in A$ such that ${\sf lin}_{a,b} \in \Gamma$, or
   \item $\exists\,\,a_0\ne b_0,\dots, a_{2k}\ne b_{2k}\in A$ such that
   $\begin{picture}(10,10)\end{picture}_{b_0}^{a_0}\begin{picture}(10,10)\put(0,10){\line(1,-1){10}}\put(0,0){\line(1,1){10}}\put(0,10){\line(1,0){10}}\end{picture}_{b_1}^{a_1},\dots, \begin{picture}(10,10)\end{picture}_{b_{2k-1}}^{a_{2k-1}}\begin{picture}(10,10)\put(0,10){\line(1,-1){10}}\put(0,0){\line(1,1){10}}\put(0,10){\line(1,0){10}}\end{picture}_{b_{2k}}^{a_{2k}}, \begin{picture}(10,10)\end{picture}_{b_{2k}}^{a_{2k}}\begin{picture}(10,10)\put(0,10){\line(1,-1){10}}\put(0,0){\line(1,1){10}}\put(0,10){\line(1,0){10}}\end{picture}_{b_0}^{a_0}\in \Gamma$, or
   \item $\Gamma$ is tractable.
\end{itemize}
\end{thm}

\proof[Proof.] Consider a functional clone $Pol\left(\Gamma\right)$ and an algebra $\left(A, Pol\left(\Gamma\right)\right)$.
Recall that the necessary local conditions are equivalent to
requiring a conservative algebra to have only tractable 2-element subalgebras.
It is obvious from the proof of Lemma \ref{L:Post} that a conservative algebra $F$ with domain set $\left\{a,b\right\}$ is NP-hard
if and only if $\begin{picture}(10,10)\end{picture}_{b}^{a}\begin{picture}(10,10)\put(0,10){\line(1,-1){10}}\put(0,0){\line(1,1){10}}\put(0,10){\line(1,0){10}}\end{picture}_{b}^{a} \in Inv\left(F\right)$ or $\begin{picture}(10,10)\end{picture}_{a}^{b}\begin{picture}(10,10)\put(0,10){\line(1,-1){10}}\put(0,0){\line(1,1){10}}\put(0,10){\line(1,0){10}}\end{picture}_{a}^{b} \in Inv\left(F\right)$ or ${\sf lin}_{a,b} \in Inv\left(F\right)$. Otherwise, it is tractable.
Therefore, the necessary local conditions for $Pol\left(\Gamma\right)$ are equivalent that $\forall\,\,a\ne b\in A$, $\begin{picture}(10,10)\end{picture}_{b}^{a}\begin{picture}(10,10)\put(0,10){\line(1,-1){10}}\put(0,0){\line(1,1){10}}\put(0,10){\line(1,0){10}}\end{picture}_{b}^{a} \notin \Gamma$ and ${\sf lin}_{a,b} \notin \Gamma$.

Suppose $\Gamma$ has the last two properties, i.e. $Pol\left(\Gamma\right)$ satisfies the necessary local conditions. As is easily seen from the proof of the NP-hard case of Theorem \ref{T:7}, $\Gamma$ is NP-hard only if it contains an odd number of predicates $\begin{picture}(10,10)\end{picture}_{b_0}^{a_0}\begin{picture}(10,10)\put(0,10){\line(1,-1){10}}\put(0,0){\line(1,1){10}}\put(0,10){\line(1,0){10}}\end{picture}_{b_1}^{a_1},\dots, \begin{picture}(10,10)\end{picture}_{b_{2k-1}}^{a_{2k-1}}\begin{picture}(10,10)\put(0,10){\line(1,-1){10}}\put(0,0){\line(1,1){10}}\put(0,10){\line(1,0){10}}\end{picture}_{b_{2k}}^{a_{2k}}, \begin{picture}(10,10)\end{picture}_{b_{2k}}^{a_{2k}}\begin{picture}(10,10)\put(0,10){\line(1,-1){10}}\put(0,0){\line(1,1){10}}\put(0,10){\line(1,0){10}}\end{picture}_{b_0}^{a_0}$. If we assume that for any $a_0\ne b_0,\dots, a_{2k}\ne b_{2k}\in A$ this system of predicates is not contained in $\Gamma$, then $\Gamma$ is tractable.
\qed

\section{Related work and open problems}
 $MinHom$ can be viewed as a problem that fits the VCSP (Valued CSP) framework by
 \cite{cohen}. By a valued predicate of
 arity $m$ over a domain $D$, we mean a function $p:D^m \rightarrow
 \mathbb{N}\cup \{\infty\}$.
 Informally, if $\Gamma$ is a finite set of valued predicates over a
 finite domain $D$, then an instance of $VCSP(\Gamma)$ is a set of variables
 together with specified subsets of variables restricted by valued
 predicates from $\Gamma$. Any assignment to variables can be considered
 a solution and the measure of this solution is the sum of the values that the
 valued predicates take under the assignments of the specified
 subsets of variables. The problem is to minimize this measure. It is
 widely believed that a dichotomy conjecture holds for $VCSP(\Gamma)$, too.

 Our dichotomy
 result for $MinHom$ encourages us to consider
 generalizations that belong to this framework.

1. Suppose we are given a constraint language $\Gamma$ and a finite set of unary functions $F \subseteq \{f:D \rightarrow \mathbb{N}\}$. Let $MinHom_F(\Gamma)$ denote a minimization problem which is defined completely analogously to $MinHom(\Gamma)$ except that we are restricted to minimizing functionals of the following form: $\sum\limits_{i = 1}^n \sum\limits_{f \in F}{w_{if}f\left( x_{i} \right)}$. A complete classification of the complexity of this problem is an open question.

 2. Suppose we have a finite valued constraint language $\Gamma$, i.e. a
 set of valued predicates over some finite domain set. If $\Gamma$
 contains all unary valued predicates, we call $VCSP(\Gamma)$ a
 conservative $VCSP$. This name is motivated by the fact that in this case the
 multimorphisms (which is a generalization of polymorphisms for valued constraint
 languages \cite{cohen}) of $\Gamma$ must consist of
 conservative functions. Since there is a well-known dichotomy for
 conservative CSPs \cite{bulatov}, we suspect that there is a dichotomy for
 conservative $VCSPs$.

 3. $MinHom$ has (just as CSP) a homomorphism formulation. If we restrict
 ourselves to relational structures given by digraphs, we arrive at the
 following problem which we call digraph $MinHom$: given digraphs $S,H$
 and weights $w_{ij}, i\in S, j\in H$, find a homomorphism $h:S\rightarrow
 H$ that minimizes the sum $\sum\limits_{s \in S} {w_{sh\left( s
 \right)}}$. Suppose we have sets of digraphs ${\mathbb G_1}, {\mathbb
 G_2}$. Then, $MinHom({\mathbb G_1},{\mathbb G_2})$ denotes the digraph
 $MinHom$ problem when the first digraph is from ${\mathbb G_1}$ and the
 second is from ${\mathbb G_2}$. In this case, $MinHom(\{H\}, All)$ is always polynomially tractable and
 $MinHom(All,\{H\})$ coincides with
 $MinHom(\{H\})$ which is characterized in this paper. Another
 characterization based on digraph theory was announced during the
 preparation of the camera-ready version of this paper \cite{hell}. We
 believe that this approach could be fruitful for characterizing the
 complexity of $MinHom({\mathbb G},{\mathbb G})$: for example, is there a dichotomy
 for $MinHom({\mathbb G},{\mathbb G})$?

\section*{Acknowledgement}

The author wishes to acknowledge fruitful discussions with Peter Jonsson and Andrei Bulatov.

%% in general the use of bibtex is encouraged

\end{document}